\documentclass[10pt,journal,letterpaper,compsoc]{IEEEtran}
\ifCLASSOPTIONcompsoc
\else
\fi
\ifCLASSINFOpdf
\else
\fi
\usepackage{times}
\usepackage{epsfig}
\usepackage{graphicx}
\usepackage{amsmath}
\usepackage{amssymb}
\usepackage[usenames, dvipsnames]{color}

\usepackage{caption}
\usepackage[table,xcdraw]{xcolor}
\usepackage{color}
\usepackage[normalem]{ulem}
\usepackage{float}
\usepackage{threeparttable}
\usepackage{multirow}
\usepackage{booktabs}
\usepackage[table]{xcolor}
\usepackage{arydshln}
\usepackage{threeparttable}
\usepackage{colortbl}
\usepackage[numbers]{natbib}
\usepackage{bm}
\usepackage{subcaption}
\usepackage{natbib}
\usepackage{array}
\usepackage[table]{xcolor}
\usepackage{colortbl}
\usepackage{siunitx}
\usepackage{url}

\usepackage[table]{xcolor}
\usepackage{colortbl}

\usepackage[pagebackref=false,breaklinks=true,colorlinks,bookmarks=false]{hyperref}

\begin{document}
%
\title{Graph Neural Network and Spatiotemporal Transformer Attention for 3D Video Object Detection from Point Clouds}
%
%
%
%

\author{Junbo Yin, Jianbing Shen,~\IEEEmembership{Senior Member,~IEEE},
Xin Gao, David Crandall, and Ruigang Yang
\IEEEcompsocitemizethanks{
\IEEEcompsocthanksitem J. Yin is with School of Computer Science, Beijing Institute of Technology, Beijing, China. \protect 
(Email: yinjunbo@bit.edu.cn)
\IEEEcompsocthanksitem J. Shen is with Inception Institute of Artificial Intelligence, Abu Dhabi, UAE.
(Email: shenjianbingcg@gmail.com)
\IEEEcompsocthanksitem  X. Gao is with Computer, Electrical, and Mathematical Sciences and Engineering (CEMSE) Division,
King Abdullah University of Science and Technology (KAUST), Thuwal, Saudi Arabia.
\IEEEcompsocthanksitem D. Crandall is with the School of Informatics, Computing, and Engineering,
Indiana University, Bloomington, IN 47408.
(Email: djcran@indiana.edu)
\IEEEcompsocthanksitem R. Yang is with the University of Kentucky, Lexington, KY 40507.
(Email: ryang@cs.uky.edu)
%
\IEEEcompsocthanksitem A preliminary version of this work has appeared in CVPR 2020~\cite{yin2020lidar}.
\IEEEcompsocthanksitem Corresponding author: \textit{Jianbing Shen}
}
\thanks{}}

%

\markboth{IEEE Transactions on Pattern Analysis and Machine Intelligence} 
{Shell \MakeLowercase{\textit{et al.}}: Bare Demo of IEEEtran.cls for Computer Society Journals}
%


\IEEEcompsoctitleabstractindextext{%
\begin{abstract}
Previous works for LiDAR-based 3D object detection mainly focus on the single-frame paradigm. In this paper, we propose to detect 3D objects by exploiting temporal information in multiple frames, \textit{i.e.}, the point cloud \textit{videos}. We empirically categorize the temporal information into short-term and long-term patterns. To encode the short-term data, we present a Grid Message Passing Network (GMPNet), which considers each grid (\textit{i.e.}, the grouped points) as a node and constructs a $k$-NN graph with the neighbor grids. To update features for a grid, GMPNet iteratively collects information from its neighbors, thus mining the motion cues in grids from nearby frames. To further aggregate the long-term frames, we propose an Attentive Spatiotemporal Transformer GRU (AST-GRU), which contains a Spatial Transformer Attention (STA) module and a Temporal Transformer Attention (TTA) module. STA and TTA enhance the vanilla GRU to focus on small objects and better align the moving objects. Our overall framework supports both online and offline video object detection in point clouds. We implement our algorithm based on prevalent anchor-based and anchor-free detectors. The evaluation results on the challenging nuScenes benchmark show the superior performance of our method, achieving the 1st on the leaderboard without any bells and whistles, by the time the paper is submitted.
\end{abstract}

}

\maketitle

\IEEEdisplaynotcompsoctitleabstractindextext

%
\IEEEpeerreviewmaketitle

\section{Introduction}
\begin{figure}
\begin{center}
\includegraphics[width=0.498\textwidth]{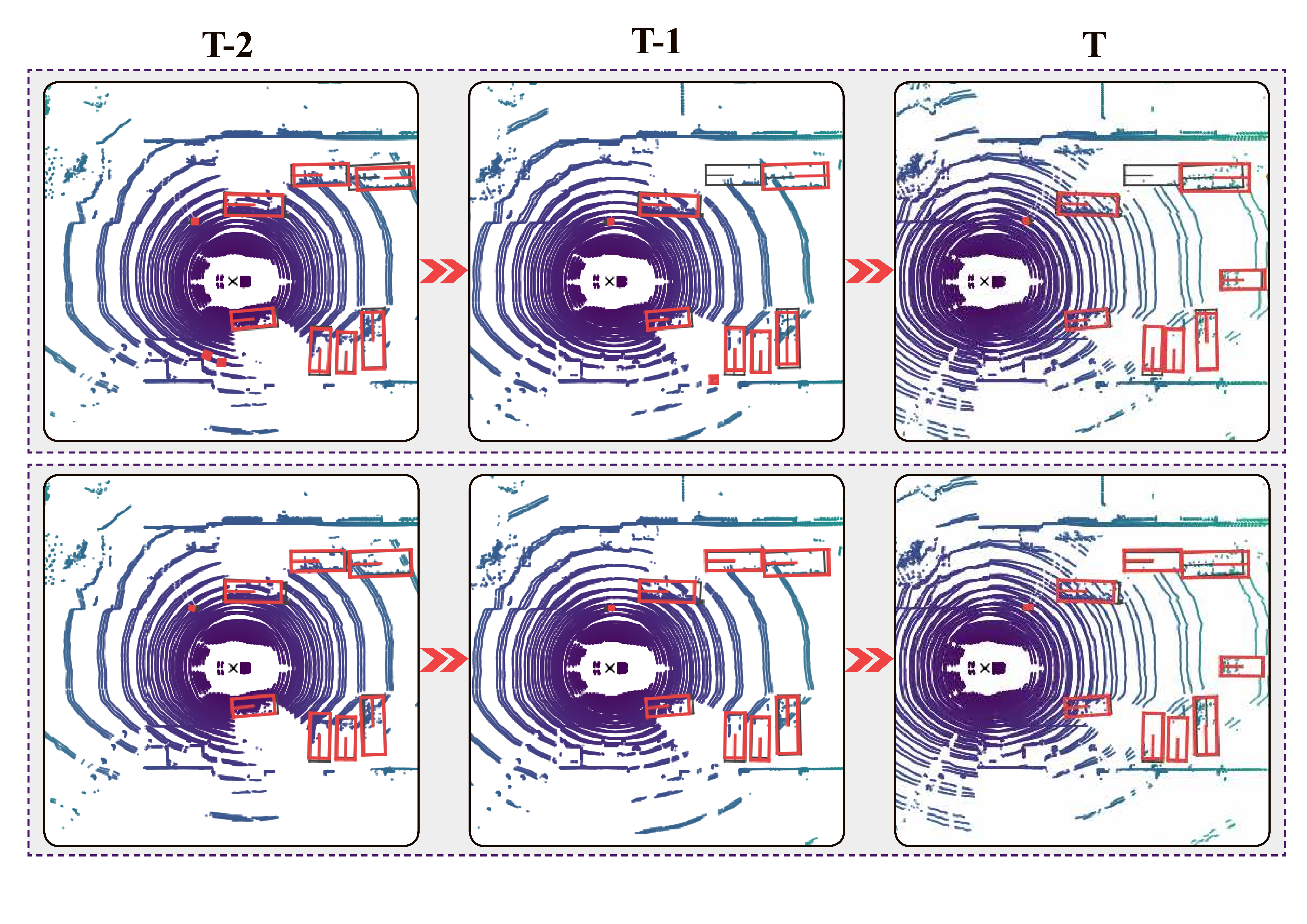}
\end{center}
\caption{\small {\textbf{A representative challenging case in autonomous driving scenarios.} A typical single-frame 3D object detector,
\textit{eg.} \cite{lang2019pointpillars}, leads to false-negative (FN) results due to occlusion  (top row). In contrast, our 3D video object detector could address this (bottom row). The red and grey boxes denote the predictions and ground-truths, respectively.}
}
\label{fig:video}
\end{figure}

\IEEEPARstart{T}{he} past several years have witnessed an explosion of interests in 3D object detection due to its vital role in autonomous driving perception. Meanwhile, the LiDAR sensor is becoming an indispensable instrument in 3D perception for its capacity of providing accurate depth information in intricate scenarios such as dynamic traffic environments and adverse weather conditions. The majority of 3D object detectors are dedicated to detecting in \textit{single-frame} LiDAR point clouds, \textit{i.e.}, predicting oriented 3D bounding boxes via frame-by-frame paradigm. However, little work has been devoted to detecting in {multi-frame} point clouds sequences, \textit{i.e.}, point cloud \textit{videos}. In the newly released nuScenes~\cite{caesar2020nuscenes} dataset, a point cloud video is defined by a driving scene containing around 400 point cloud frames captured within 20 seconds. In general, point cloud videos are readily available in practical applications and can provide rich spatiotemporal coherence. For example, object point clouds from a single frame may be truncated or sparse due to occlusions or long-distance sampling, while other frames could contain complementary information for recognizing the object. Therefore, a \textit{single-frame} 3D object detector suffers from deteriorated performance in comparison with a 3D \textit{video} object detector (See Fig.~\ref{fig:video}).

Existing 3D video object detectors typically exploit the temporal information in a straightforward way, \textit{i.e.}, directly concatenating the point clouds from other frames to a reference frame and then performing detection in the reference frame~\cite{lang2019pointpillars,hu2020you,yin2020center}. This converts the video object detection task to a single-frame object detection task. However, such a simple approach has several limitations. First, in driving scenarios, there will be serious motion blur that will result in inaccurate detection results. One solution might be to apply ego-motion transformation for these frames. Though it can alleviate the ego-motion, it is not capable of remedying the motion blur caused by moving objects. Second, this concatenation-based approach usually simply encodes the temporal information by an additional point cloud channel such as timestamp, which ignores the rich feature relations among different frames. Third, as the label information is only provided in the reference frame, such methods will suffer more information loss when leveraging long-term point cloud data. Apart from the concatenation-based approach, other works~\cite{luo2018fast} apply temporal 3D convolutional networks on point cloud videos. Nevertheless, they will encounter temporal information collapse when aggregating features over multiple frames~\cite{tran2018closer}.

How to effectively exploit the temporal information in point cloud videos remains an open challenge. We argue that there are two forms of temporal information, \textit{i.e.}, short-term and long-term patterns. Taking the nuScenes~\cite{caesar2020nuscenes} dataset for example, the short-term pattern is defined by point cloud sequences captured within 0.5 seconds, including around 10 frames. As for the long-term pattern, we refer to the point cloud frames collected in 1 to 2 seconds involving dozens of frames. In this work, we aim to present a more effective and general 3D video object detection framework by enhancing prevalent 3D object detectors with both the short-term and long-term temporal cues. In particular, the long-term point clouds are first divided into several short-term ones. We separately encode each short-term data with a short-term encoding module, and then adaptively fuse the output features by a long-term aggregation module.

For handling the short-term point cloud data, our short-term encoding module follows the paradigm of modern grid-based detectors~\cite{lang2019pointpillars,zhou2018voxelnet,yin2020center,yan2018second} that directly concatenate point clouds from nearby frames to a reference frame, but adopts a novel grid feature encoding strategy that models the relations of girds in nearby frames.
More precisely, these detectors typically divide the point clouds into regular girds such as voxels and pillars, with each gird containing a fixed number of point clouds. Then PointNet-like modules (\textit{e.g.}, the Voxel Feature Encoding Layer in~\cite{zhou2018voxelnet} and the Pillar Feature Network in~\cite{lang2019pointpillars}) are used to extract the grid-wise feature representation. A potential problem of these PointNet-like modules is that they only focus on the \textit{individual} grid, which ignores the relations among different grids. Intuitively, certain grids in nearby frames may encode the same object parts, which can be explored to improve the detection accuracy. To this end, we propose a graph-based network, Grid Message Passing Network (GMPNet), which iteratively propagates information over different grids. GMPNet treats each \textit{non-empty} grid as a graph node and builds a $k$-NN graph with the nearby $k$ grids. The relations among these grids as viewed edges. The core idea of GMPNet is to iteratively update the node features via the neighbors along the edges and hence mine the rich relations among different grid nodes. At each iteration, a node receives pair-wise messages from its neighbors, aggregates these messages and then updates its node features. After several iterations, messages from distant grids can be accessed. This flexibly enlarges the receptive field in a non-Euclidean manner. Compared with previous PointNet-like modules, GMPNet effectively encourages information propagation among different grids, meanwhile capturing short-term motion cues of objects.

After obtaining individual features extracted by the short-term encoding module, we then turn to Convolutional Gated Recurrent Unit networks (ConvGRU~\cite{ballas2016delving}) to further capture the dependencies over these features in our long-term aggregation module. ConvGRU has shown promising performance in the 2D video understanding field. It adaptively links temporal information over input sequences with a memory gating mechanism. In each time step, the current memory is computed by considering both the current input and the previous memory features. The updated memory is then employed to perform the present task. However, there are two potential limitations when directly applying ConvGRU on multi-frame point cloud sequences. First, modern grid-based detectors tend to transform the point cloud features into the bird's eye view, and the resultant object resolution is much smaller than that in 2D images. For example, with a grid size of $0.25^2~m^2$ and stride of $8$, objects such as pedestrians and vehicles just occupy around 1 to 3 pixels in the output feature maps. Therefore, when computing the current memory features, the background noise will be inevitably accumulated and decrease detection performance. Second, the spatial features of the current input and the previous memory are not well aligned across frames. Though we could use the ego-pose information to align the static objects over multiple point cloud frames, the moving objects still incur motion blur that impairs the quality when updating the memory.

To address these challenges, we present Attentive Spatiotemporal Transformer GRU (AST-GRU), which enhances the vanilla ConvGRU with a Spatial Transformer Attention (STA) module and a Temporal Transformer Attention (TTA) module. In particular, the STA module is derived from~\cite{vaswani2017attention, wang2018non} and is devised to attend the small objects with the spatial context information. It acts as an intra-attention mechanism, where both the keys and queries reside in the input features. STA views each pixel as a query and the neighbors as keys. By attending the query with the context keys, STA can better tell foreground pixels from the background ones, and thus emphasize the meaningful objects in current memory features. Furthermore, we also describe a TTA module motivated by ~\cite{zhu2019deformable,zhu2019empirical}, which is employed to align the moving objects. It is composed of modified deformable convolutional layers and behaves as an inter-attention that operates on both the input and previous memory features. TTA treats each pixel in the previous memory as a query and decides the keys by integrating the motion information, thus capturing more cues of moving objects. In contrast to the vanilla ConvGRU, the presented AST-GRU effectively improves the quality of the current memory features and leads to better detection accuracy, which is verified on both prevalent anchor-based and anchor-free detectors. Moreover, this also introduces a general methodology that accounts for both online and offline 3D video object detection by flexibly deciding the inputs,~\textit{e.g.}, using previous or later frames in a point cloud video. In the offline mode, we further augment our AST-GRU with a bi-directional feature learning mechanism, which achieves better results in comparison to the online mode.

To summarize, the main contributions of this work are as follows:
\begin{itemize}
\item A general point cloud-based 3D video object detection framework is proposed by leveraging both short-term and long-term point clouds information. The proposed framework can easily integrate prevalent 3D object detectors in both online and offline modes, which provides new insights to the community.
\item We present a novel graph neural network, named GMPNet, to encode short-term point clouds. GMPNet can flexibly mine the relations among different grids in nearby frames and thus capture the motion cues.
\item To further model the long-term point clouds, an AST-GRU module is introduced to equip the conventional ConvGRU with an attentive memory mechanism, where a Spatial Transformer Attention (STA) and a Temporal Transformer Attention (TTA) are devised to mine the spatiotemporal coherence of long-term point clouds.
\item We build the proposed 3D video object detection framework upon both anchor-based and anchor-free 3D object detectors. Extensive evaluations on the large-scale nuScenes benchmark show that our model outperforms all the state-of-the-art approaches on the leaderboard.
\end{itemize}

This work significantly extends our preliminary conference paper~\cite{yin2020lidar} and the improvements are multi-fold. Our previous work~\cite{yin2020lidar} only supports 3D object detection in the online mode with an anchor-based detector. In this work, we first provide a more general framework that works with both anchor-based and anchor-free detection settings. In this regard, our approach can be easily incorporated with leading 3D object detectors. Second, we extend our method to both online and offline detection modes, which can benefit more applications with improved accuracy, such as acting as annotation tools. Third, this paper provides a more in-depth discussion on the algorithm with more details including its motivation, technical preliminary, network architecture and implementation. Fourth, extensive ablation studies are conducted to thoroughly and rigorously assess the broad effectiveness of our model. Last but not least, we empirically observe that the proposed model outperforms all the algorithms on the nuScenes leaderboard without any bells and whistles.

\begin{figure*}[t]
  \centering
      \includegraphics[width=0.99\linewidth]{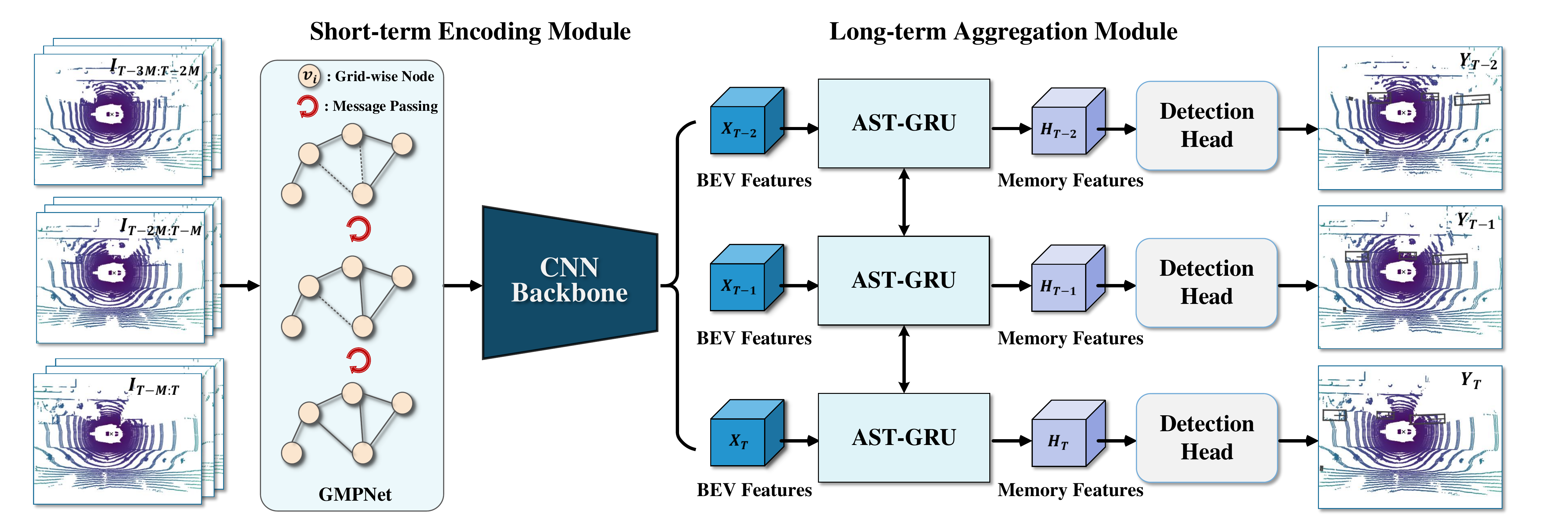}
\caption{\small \textbf{Schematic of our point cloud-based 3D video object detection framework.} {The input $T\times M$ frames are first divided into $T$ groups of short-term data with each merging $M$ frames.} Then, the short-term encoding module extracts the BEV features for each short-term data with a Grid Message Passing Network (GMPNet) followed by a CNN backbone. Afterwards, the long-term aggregation module further captures the dependencies in these short-term features with an Attentive Spatiotemporal Transformer GRU (AST-GRU). Finally, the detection head receives the enhanced memory features and produces the detection results.}
\label{fig:framework}
\end{figure*}


\section{Related Work}

\noindent\textbf{3D Object Detection.} A considerable amount of efforts have been made into  3D object detection in recent years for its crucial role in autonomous driving. Existing approaches for 3D object detection can be roughly grouped into three categories.  \textit{(1) 2D image-based methods} typically perform detection from monocular~\cite{chen2020monopair, ma2019accurate,ku2019monocular} or stereo images~\cite{li2019stereo,chen2020dsgn,wang2019pseudo}. These approaches often experience great difficulty in getting promising performance due to the information loss of depth. They often turn to geometric priors~\cite{chen2016monocular} or integrate an additional module to estimate depth~\cite{ma2019accurate} or disparity~\cite{li2019stereo}.  It is worth noting that Wang~\textit{et al.}~\cite{wang2019pseudo} presented a novel framework by first converting image-based depth maps to pseudo point clouds and then applying the off-the-shelf LiDAR-based detectors. \textit{(2) 3D point cloud-based methods} can leverage the LiDAR sensor to access the accurate depth information, and are less sensitive to different illumination and weather conditions. Grid- and Point-based methods are the main ways to process the point clouds. The common solution of grid-based methods~\cite{zhou2018voxelnet,yang2018pixor,lang2019pointpillars,yin2020lidar} is to first discretize the raw point clouds into regular girds (\textit{e.g.}, voxels~\cite{yan2018second} or pillars~\cite{lang2019pointpillars}). Then 3D or 2D convolutional networks can be readily applied to extract the features. In practice, grid-based approaches are much more efficient than point-based methods, but may suffer from information loss in the quantification process. Point-based methods~\cite{shi2019pointrcnn, yang2019std, chen2019fast} typically extract features and predict proposals from point-level with backbones like PointNet++~\cite{qi2017pointnet}. They are prone to get better performance than the grid-based methods, but are limited to computational efficiency and memory footprint when the number of point clouds increases. Recently, Shi~\textit{et al.}~\cite{shi2020pv} proposed PV-RCNN detector that combines the merits of both the voxel- and point-based approaches and shows better performance. \textit{(3)  Fusion-based methods} exploit both the camera and LiDAR sensors to capture complementary information. MV3D~\cite{chen2017multi} is the seminal work of this family. It takes LiDAR bird's eye view and front view as well as images as inputs, and combines the proposal-wise features of each view with a deep fusion network. Later, 3D-CVF~\cite{yoo20203d} proposed to combine point clouds and images from N multi-view cameras with a cross-view spatial feature fusion strategy. However, the multi-sensor fusion system may not work robustly due to signal synchronization problems. In this work, we focus on the LiDAR-only grid-based approaches since they are more prevalent in current autonomous driving applications.

 \noindent\textbf{Spatiotemporal Models in Point Clouds.}
Different from the aforementioned works that only perform frame-by-frame detections, we aim to address the multiple-frame 3D object detection by exploiting the temporal information. Only a few works have explored the spatiotemporal model in point clouds. Luo~\textit{et al.}~\cite{luo2018fast} utilized temporal 3D ConvNet to aggregate the multi-frame point clouds. It performs multiple tasks simultaneously including detection, tracking and motion prediction, but leads to a considerable amount of parameters for operating on 4D tensors. {Liang~\textit{et al.}~\cite{liang2020pnp} also addressed the framework of joint detection, tracking and motion prediction for improving efficiency and accuracy.} Later, Choy~\textit{et al.}~\cite{choy20194d} improved the convolutional layers in ~\cite{luo2018fast} with sparse operations. However, it encounters the feature collapse issue when downsampling the features in the temporal domain, and fails to leverage full label information in all frames. More recently, Huang~\textit{et al.}~\cite{huang2020lstm} proposed a contemporary work with ours that focuses on a point-based recurrent gating mechanism. It adopts a Sparse Conv U-Net to extract the spatial features of each frame, and then applies LSTM on points with high semantic scores to fuse information across frames and save computation. Due to the restriction of the backbone, it is not able to be adapted to the state-of-the-art single-frame detectors. On the contrary, our work can be easily integrated into prevalent grid-based 3D object detectors~\cite{lang2019pointpillars,yin2020center} by processing the spatiotemporal information in point cloud videos.
It is worth mentioning that there are some 2D video object detection works~\cite{chen2020mega, feifei2020vid} also explore the attention modules for modeling temporal information. However, for 3D point cloud data with bird's eye view, there are typically more moving objects containing sparse points with shorter lifespans. {Our work aims to address these new challenges in 3D video object detection by aggregating point features in consecutive frames.} 

\noindent\textbf{Graph Neural Networks.}
The concept of Graph Neural Networks (GNNs) was first proposed by Scarselli~\textit{et al.}~\cite{scarselli2008graph}. They extended the Recursive Neural Networks (RNN) and used GNNs to directly process the graph-structured data. It encodes the underlying relationships among the nodes of the graph and mines the rich information in the data processing step. Afterwards, several variants have been developed to improve the representation capability of the original GNNs. Here, we categorize them into two classes according to the type of the information propagation step: convolution and gate mechanism. Graph Convolutional Networks (GCNs)~\cite{bruna2014spectral,henaff2015deep,defferrard2016convolutional,hammond2011wavelets,niepert2016learning} belongs to the former group that generalize convolution to the graph data and update nodes via stacks of graph convolutional layers. GCNs typically compute in the spectral domain with graph Fourier transformation. Recently, various applications~\cite{hamilton2017inductive,schlichtkrull2018modeling,gao2019graph} have been explored by GCNs, which achieve promising performance. The latter group~\cite{li2016gated, kearnes2016molecular, zayats2018conversation, peng2017cross} aimed to adopt recurrent gating units to propagate and update information across nodes. For instance, Li~\textit{et al.}~\cite{li2016gated} exploited the Gated Recurrent Unit (GRU) to describe the node state by aggregating messages from neighbors. After that, Gilmer~\textit{et al.}~\cite{gilmer2017neural} proposed a generalized framework that formulates the graph computation as a parameterized Message Passing Neural Network (MPNN).  MPNN has been well demonstrated the ability in various tasks that involve graph data~\cite{wang2019zero,qi2018learning,si2018skeleton}. Our GMPNet is also inspired by MPNN that encodes the features of short-term point clouds and mines the motion cues among the grid-wise nodes by iterative message passing.

\section{The Proposed 3D Video Object Detection Framework}
In this section, we first present the overall pipeline of our method in \S\ref{subsec:overview}. Then, we briefly review the general formulations of Message Passing Neural Network~\cite{gilmer2017neural} and ConvGRU~\cite{ballas2016delving} in \S\ref{subsec:MPNN} and \S\ref{subsec:convGRU}, respectively, since our work mainly builds upon these works. Afterwards, the crucial designs of Grid Message Passing Network (GMPNet) and Attentive Spatiotemporal Transformer GRU (AST-GRU) are described in \S\ref{subsec:GMPNet} and \S\ref{subsec:ASTGRU}, respectively. Notice that the main part of this paper follows the online object detection pipeline. While in \S\ref{subsec:offline}, an offline pipeline is also presented by learning with bi-directional ConvGRU, which further improves the detection performance. Finally, we provide more detailed information on our network architecture in \S\ref{subsec:detail}.

\subsection{Overview}
\label{subsec:overview}
 As illustrated in Fig.~\ref{fig:framework}, our framework consists of a short-term encoding module and a long-term aggregation module. Assuming that the input long-term point cloud sequence $\{\bm{I}_t\}_{t=1}^{T\times M}$ contains $T\times M$ frames in total, we first divide it into multiple short-term clip  $\{\bm{I}_t\}_{t=1}^{T}$, with each $\bm{I}_t$ containing concatenated point clouds within $M$ nearby frames (\textit{eg.}, $M=10$ in nuScenes).  Then, LiDAR pose information is leveraged to align point clouds over frames in $\{\bm{I}_t\}_{t=1}^{T}$ to eliminate the influence of ego-motion. Next, each $\bm{I}_t$ is quantized into evenly distributed grids, and forwarded to the short-term encoding module to extract the bird's eye view features $\{\bm{X}_t\}_{t=1}^{T}$. Here, GMPNet is applied before the 2D CNN backbone to capture the relations among grids. After that, in the long-term aggregation module, AST-GRU further learns the long-term dependencies of the sequential features $\{\bm{X}_t\}_{t=1}^{T}$. In particular, at each time step $t$, the unit receives current input features $\bm{X}_t$ as well as the memory features $\bm{H}_{t-1}$, and produces the updated memory features $\bm{H}_{t}$ with an attentive gating mechanism.  In this way, $\bm{H}_{t}$ preserves the information of both previous and current frames, which facilitates the subsequent object detection task. Finally, detection heads such as classification and regression network branches can be exploited on $\bm{H}_{t}$ to obtain the final detections $\{\bm{Y}_t\}_{t=1}^{T}$.

 For training the 3D video object detector, both the anchor-based~\cite{lang2019pointpillars,yan2018second} and anchor-free~\cite{yin2020center,chen2020object} loss functions can be applied in our framework. Furthermore, our framework works in both the online and offline inference modes. In the online mode, only previous frames are employed to help the detection of the current frame. In the offline mode, both previous and later frames are viewed as supporting frames to produce detections in the current frame, which further enhances the detection performance.

\subsection{Message Passing Neural Network}
\label{subsec:MPNN}
Message Passing Neural Network (MPNN)~\cite{gilmer2017neural} is a general formulation, which unifies various promising neural networks that operate on graph-structured data~\cite{duvenaud2015convolutional,li2016gated,battaglia2016interaction,kearnes2016molecular}. Formally, it defines a directed graph $\mathcal{G}=(\mathcal{V},\mathcal{E})$, with node $v_i \in \mathcal{V}$ and edge ${e}_{i, j}\in\mathcal{E}$. The core of MPNN is to iteratively pass messages between different nodes and mine the diverse relations of them. At each time step $t$, let $\bm{h}_i^{t}$ denote the state features of node $v_i$,  and  $\bm{e}_{j, i}^{t}$ represents edge features of ${e}_{i, j}$ that describes the information passed from node $v_j$ to $v_i$. MPNN aggregates information for node $v_i$ with the neighbors  ${v_j}\in{\bm{\Omega}_{v_i}}$, and infers its updated state features $\bm{h}_i^{t+1}$ based on the received messages. It runs for $K$ time steps and thus captures the long range interactions among the nodes.

More specifically, MPNN includes a message function $M(\cdot)$ and a node update function $U(\cdot)$. At each time step $t$, $M(\cdot)$ summarizes the message $\bm{m}_{j, i}^{t+1}$ passed from the neighbor node ${v_j}\in{\bm{\Omega}_{v_i}}$ to $v_i$, then obtains the aggregated message features $\bm{m}_{i}^{t+1}$. Notice that the message features $\bm{m}_{j, i}^{t+1}$ is computed by considering both the node state features and edge features, which is denoted as follows:
\begin{equation}
\begin{aligned}
\label{eq:message}
\bm{m}_{i}^{t+1} &= \mathop{\sum}_{j\in\bm{\Omega}_i}\bm{m}_{j, i}^{t+1}  \\
&=\mathop{\sum}_{j\in\bm{\Omega}_i}M(\bm{h}_i^t,\bm{h}_j^t,\bm{e}_{j, i}^t).
\end{aligned}
\end{equation}
Then, according to the collected message $\bm{m}_{i}^{t+1}$, the update functions $U(\cdot)$ refines the previous state features $\bm{h}_i^t$ for node $v_i$, and produces the updated state features $\bm{h}_i^{t+1}$:
\begin{equation}
\begin{aligned}
\label{eq:update_org}
\bm{h}_{i}^{t+1} = U(\bm{h}_i^t, \bm{m}_{i}^{t+1}).
\end{aligned}
\end{equation}

In MPNN, both $M(\cdot)$ and $U(\cdot)$ are parameterized by weight-sharing neural networks and all the operations can be learnt with gradient-based optimization. After one time step, a node accesses information from its neighbor nodes. After $K$ time step, the information from long-range nodes can be obtained. In this work, we extend MPNN to the context of LiDAR point clouds by treating grids as nodes. Our proposed GMPNet can encode the grid-wise features by mining the spatiotemporal relations in short-term point clouds.

\subsection{ConvGRU Network}
\label{subsec:convGRU}
Gated Recurrent Unit (GRU) model~\cite{cho2014learning} is a streamlined variant of Recurrent Neural Network (RNN)~\cite{bahdanau2014neural,srivastava2015unsupervised,chung2014empirical}, which is originally devised for machine translation and video understanding by capturing the dependencies of input sequences. It simplifies the computation of RNN and achieves comparable performance. Later, Ballas~\textit{et al.}~\cite{ballas2016delving} proposed convolutional GRU (ConvGRU), which utilizes convolutional layers to replace the fully-connected ones in the original GRU. It not only preserves the better spatial resolution of the input features, but also largely reduces the number of parameters. ConvGRU has shown promising results on many vision tasks~\cite{tran2018closer, liu2018mobile, feichtenhofer2017spatiotemporal, gan2017semantic}. Basically, ConvGRU contains an update gate $\bm{z}_t$, a reset gate $\bm{r}_t$, a candidate memory $\tilde{\bm{H}}_t$ and a current memory $\bm{H}_t$. At each time step, it computes the current memory $\bm{H}_t$ (or the hidden state) according to the previous memory $\bm{H}_{t-1}$ and the current input $\bm{X}_t$, which is denoted by the following equations:
\begin{equation}
\label{eq:gru}
\begin{split}
&\bm{z}_{t} = \sigma(\bm{W}_{z}*\bm{X}_{t}+\bm{U}_{z}*\bm{H}_{t-1}),\\
&\bm{r}_{t} = \sigma(\bm{W}_{r}*\bm{X}_{t}+\bm{U}_{r}*\bm{H}_{t-1}),\\
&\tilde{\bm{H}}_{t} = \tanh(\bm{W}*\bm{X}_{t}+\bm{U}*(\bm{r}_{t}\circ\bm{H}_{t-1})),\\
&\bm{H}_{t} = (\bm{1}-\bm{z}_t)\circ\bm{H}_{t-1}+\bm{z}_t\circ\tilde{\bm{H}}_{t},
\end{split}
\end{equation}
where `*' denotes the convolution operation, `$\circ$' is the Hadamard product and $\sigma$ acts as a sigmoid activation function. $\bm{W}, \bm{W}_{z},\bm{W}_{r}$ and $\bm{U}, \bm{U}_{z},\bm{U}_{r}$ are the 2D convolutional kernels. The reset gate $\bm{r}_t$ determines how much of the past information from $\bm{H}_{t-1}$ to forget, so as to produce the candidate memory $\tilde{\bm{H}}_{t}$. For example, the information of $\tilde{\bm{H}}_{t}$ all comes from the current input $\bm{X}_t$ when $\bm{r}_{t}=0$. Besides, the update gate $\bm{z}_t$ decides the degree to which the current memory $\bm{H}_t$ accumulates the previous information form $\bm{H}_{t-1}$. Our AST-GRU significant improves the vanilla ConvGRU by integrating it with STA and TTA modules, which enforces the network to focus on meaningful objects in long-term point clouds.

\subsection{Grid Message Passing Network}
\label{subsec:GMPNet}
As introduced in~\S\ref{subsec:overview}, in the short-term encoding module, we aggregate the $K$ short-term point cloud frames by concatenating them into a single frame. To extract features on this merged frame, dominant approaches tend to quantize the point clouds into regular grids such as voxels or pillars, and then utilize modules like Voxel Feature Encoding Layer~\cite{zhou2018voxelnet} or the Pillar Feature Network~\cite{lang2019pointpillars} to encode the grids. These modules typically consider an \textit{individual} grid, which fails to capture the spatiotemporal relations between the nodes from nearby frames, as well as limiting the expressive power due to the local representation. To this end, we propose Grid Message Passing Network (GPMNet) to mine the spatiotemporal relations of the grids, which results in a non-local representation. GMPNet views each \textit{non-empty} grid as a node, and the relations between grids as edges. It iteratively passes messages between the grids and updates the grid-level representation accordingly. Besides, GPMNet also provides complementary perspectives for the subsequent CNN backbone with the non-Euclidean characteristics

Specifically, given a merged point cloud frame $\bm{I}_t$, we first uniformly discretize it into a set of grids $\mathcal{V}$. The grid can be either a voxel or a pillar, determined by the baseline detectors. Then a directed graph $\mathcal{G}=(\mathcal{V},\mathcal{E})$ is constructed, where each node $v_i \in \mathcal{V}$ represents  a non-empty grid and each edge ${e}_{j, i}\in\mathcal{E}$ holds the information passed from node $v_j$ to $v_i$. Furthermore, we define $\mathcal{G}$ as a $k$-nearest neighbor ($k$-NN) graph to save computations, which means that each node can directly access information from the $K$ spatial neighbors (also known as the first order neighbors). The goal of GMPNet is to adaptively update feature representation $\bm{h}_i$ for each grid-wise node $v_i$ by integrating information from long-range nodes, \textit{i.e.}, the higher order neighbors. Given a grid-wise node $v_i$, we first use a simplified PointNet~\cite{qi2017pointnet} module to abstract its initial state features $\bm{h}_i^0$, which map a set of points within $v_i$ to an $L$-dim vector. The simplified PointNet is composed of fully connected layers $f(\cdot)$ and a max-pooling operation:
\begin{equation}
\begin{aligned}
\label{eq:pfn}
\bm{h}_i^0 = \mathop {\text{max}}\{f{(V_i)}\}	\in\mathbb{R}^{L},
\end{aligned}
\end{equation}
where $V_i \in \mathbb{R}^{N \times D}$ denotes the grid $v_i$ with $N$ point clouds parameterized by $D$-dim representation (\textit{eg.}, the XYZ coordinates and the reflectance of LiDAR).

Next,  we elaborate the message passing and node state updating in GMPNet, following the formulation presented in~\S\ref{subsec:MPNN}. At time step $s$, $v_i$ aggregates information from its neighbors $v_j \in \bm{\Omega}_{v_i}$ according to Eq.~\ref{eq:message}. The incoming edge features $\bm{e}_{j, i}^{s}$ is defined as:
\begin{equation}
\begin{aligned}
\label{eq:edge}
\bm{e}_{j, i}^{s} = \bm{h}_j^s - \bm{h}_i^s\in\mathbb{R}^{L},
\end{aligned}
\end{equation}
 which is an asymmetric function encoding the local neighbor information. Accordingly, the message passed from $v_j$ to $v_i$ is denoted as:
\begin{equation}
\begin{aligned}
\label{eq:message1}
\bm{m}_{j, i}^{s+1} = \phi_{\theta}{([\bm{h}_i^s, \bm{e}_{j, i}^s])}\in\mathbb{R}^{L'},
\end{aligned}
\end{equation}
where $\phi_{\theta}$ is a fully connected layer, denoting the message function $M(\cdot)$ in Eq.~\ref{eq:message}. It receives the concatenation of $\bm{h}_i^s$ and $\bm{e}_{j, i}^s$, and yields an $L'$-dim feature. We then summarize the received messages from $K$ neighbors with a max-pooling operation:
\begin{equation}
\begin{aligned}
\label{eq:message2}
\bm{m}_{i}^{s+1} = \mathop {\text{max}}\limits_{j\in\bm{\Omega}_i}{\{\bm{m}_{j, i}^{s+1}\}}\in\mathbb{R}^{L'}.
\end{aligned}
\end{equation}

\begin{figure}
\begin{center}
\includegraphics[width=0.48\textwidth]{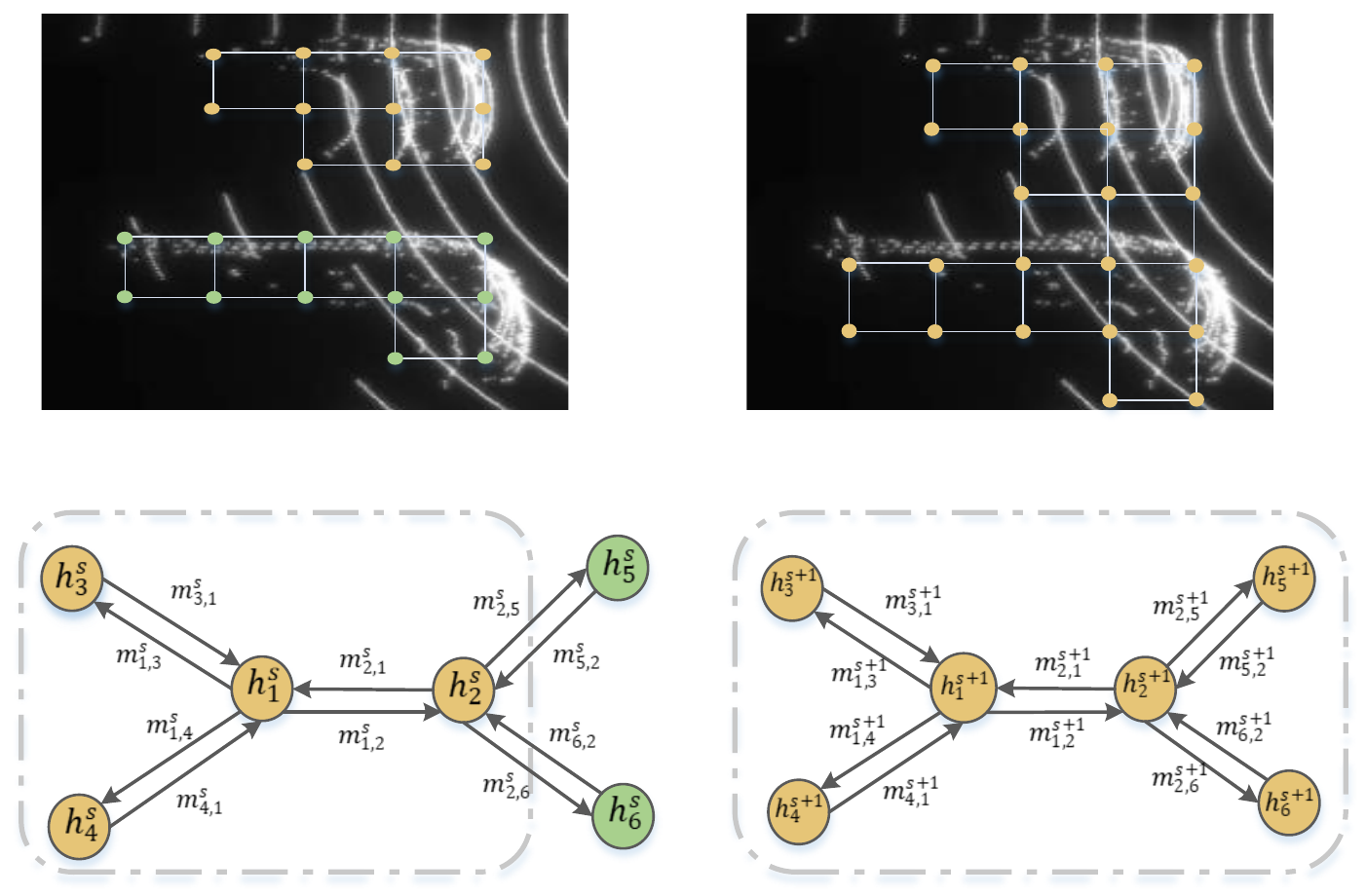}
\end{center}
\caption{\small {\textbf{Illustration of one iteration step for message propagation}, where $h_i$ is the state of node $v_i$. In step $s$, the neighbors for $h_1$ are $\{h2, h3, h4\}$ (within the gray dash line), presenting the grids of the top car. After receiving messages from the neighbors, the receptive field of $h_1$ is enlarged in step $s+1$. This indicates the relations with the bottom car are modeled after message propagation.}}
\label{fig:gnn}
\end{figure}

Afterwards, with the collected message $\bm{m}_{i}^{s+1}$, we update the state features $\bm{h}_i^s$  for node $v_i$ in terms of Eq.~\ref{eq:update_org}. Here, GRU~\cite{cho2014learning} is used as the update function $U(\cdot)$ to adaptively preserve the information in different time steps, which is:
\begin{equation}
\begin{aligned}
\bm{h}_{i}^{s+1} = {\text{GRU}}(\bm{h}_i^s, \bm{m}_{i}^{s+1})\in\mathbb{R}^{L}.
\label{eq:update}
\end{aligned}
\end{equation}
After one time step, $v_i$ contains the information from the neighbors $v_j \in \bm{\Omega}_{v_i}$. Moreover, the neighbor node $v_j$ also holds information from its own neighbors $\bm{\Omega}_{v_j}$. Therefore, $v_i$ can aggregate information from long-range nodes after a total of $S$ time steps. The message passing and node updating processes are illustrated in Fig.~\ref{fig:gnn}.

We obtain the final grid-wise feature representation $\bm{v}_{i} $ by applying another fully connected layer $\phi^{'}_{\theta}$  on $\bm{h}_{i}^{S}$:
\begin{equation}
\begin{aligned}
\label{eq:output}
\bm{v}_{i} = \phi_{\theta}^{'}{(\bm{h}_{i}^{S})}\in\mathbb{R}^{L}.
\end{aligned}
\end{equation}

After that, all the grid-wise features are then scattered back to a regular tensor $\bm{\tilde{I}}_t$, \textit{e.g.}, $\bm{\tilde{I}}_t\in\mathbb{R}^{W\times H\times L}$ with the PointPillars~\cite{lang2019pointpillars} baseline. Finally, we apply the CNN backbone in~\cite{zhou2018voxelnet} to further extract features for $\bm{\tilde{I}}_t$:
\begin{equation}
\begin{aligned}
\label{eq:2dbackbone}
\bm{X}_{t} = {F_{\text{B}}}(\bm{\tilde{I}}_t)\in\mathbb{R}^{w\times h\times c},
\end{aligned}
\end{equation}
where $F_{\text{B}}$ is the backbone network and $\bm{X}_{t}$ is the obtained short-term features of $\bm{{I}}_t$. This leads to a reduced resolution for $\bm{{I}}_t$, facilitating the subsequent long-term aggregation module. More details of the GMPNet and the CNN backbone can be found in \S\ref{subsec:detail}.

\subsection{Attentive Spatiotemporal Transformer GRU}
\label{subsec:ASTGRU}
Recall that we have divided the input point cloud sequences $\{\bm{I}_t\}_{t=1}^{T\times M}$ into multiple short-term clips $\{\bm{I}_t\}_{t=1}^{T}$, with each clip $\bm{I}_t$ including concatenated point clouds from $M$ neighbor frames. Then, in the short-term encoding module, we independently described the short-term features $\bm{X}_t$ for each $\bm{{I}}_t$. Here, we further capture the long-term dynamics over the sequential features $\{\bm{X}_t\}_{t=1}^{T}$  in the long-term aggregation module by exploiting the Attentive Spatiotemporal Transformer GRU (AST-GRU). AST-GRU is an extension of ConvGRU that adaptively mines the spatiotemporal dependencies in $\{\bm{X}_t\}_{t=1}^{T}$. In particular, it improves the vanilla ConvGRU in two ways inspired by the transformer attention mechanism~\cite{vaswani2017attention,carion2020end}. First, dominant approaches usually perform detection on bird-eye view feature maps~\cite{lang2019pointpillars,yin2020center,zhou2018voxelnet,yan2018second}. However, the interest objects are much smaller in such a view compared with the front view in the 2D images. For example, using PointPillars~\cite{lang2019pointpillars} with a voxel size of $0.25^2~m^2$, a vehicle (\textit{eg}, with size of $4m\times2m$) remains only 2 pixels in $\bm{X}_t$ after the feature extractor with common stride of 8. This causes difficulties for the recurrent unit,
because the background noise will inevitably accumulate across time in $\bm{H}_{t}$. To handle this problem, we propose a spatial transformer attention (STA) module by attending each pixel in $\bm{X}_t$ with its rich spatial context, which helps the detectors to focus on foreground objects. Second, for a recurrent unit, the received current input $\bm{X}_t$ and pervious memory $\bm{H}_{t-1}$ are not spatially aligned. Though we have aligned the static objects with the ego-pose information, the dynamic objects with large motion still lead to an inaccurate current memory $\bm{H}_{t}$. Therefore, a temporal transformer attention (TTA) module is introduced to align the moving objects in  $\bm{H}_{t}$, which exploits modified deformable convolutional networks to capture the motion cues in $\bm{H}_{t-1}$ and $\bm{X}_t$. Next, we elaborate the workflow of our STA and TTA modules.

\begin{figure}
\begin{center}
\includegraphics[width=0.49\textwidth]{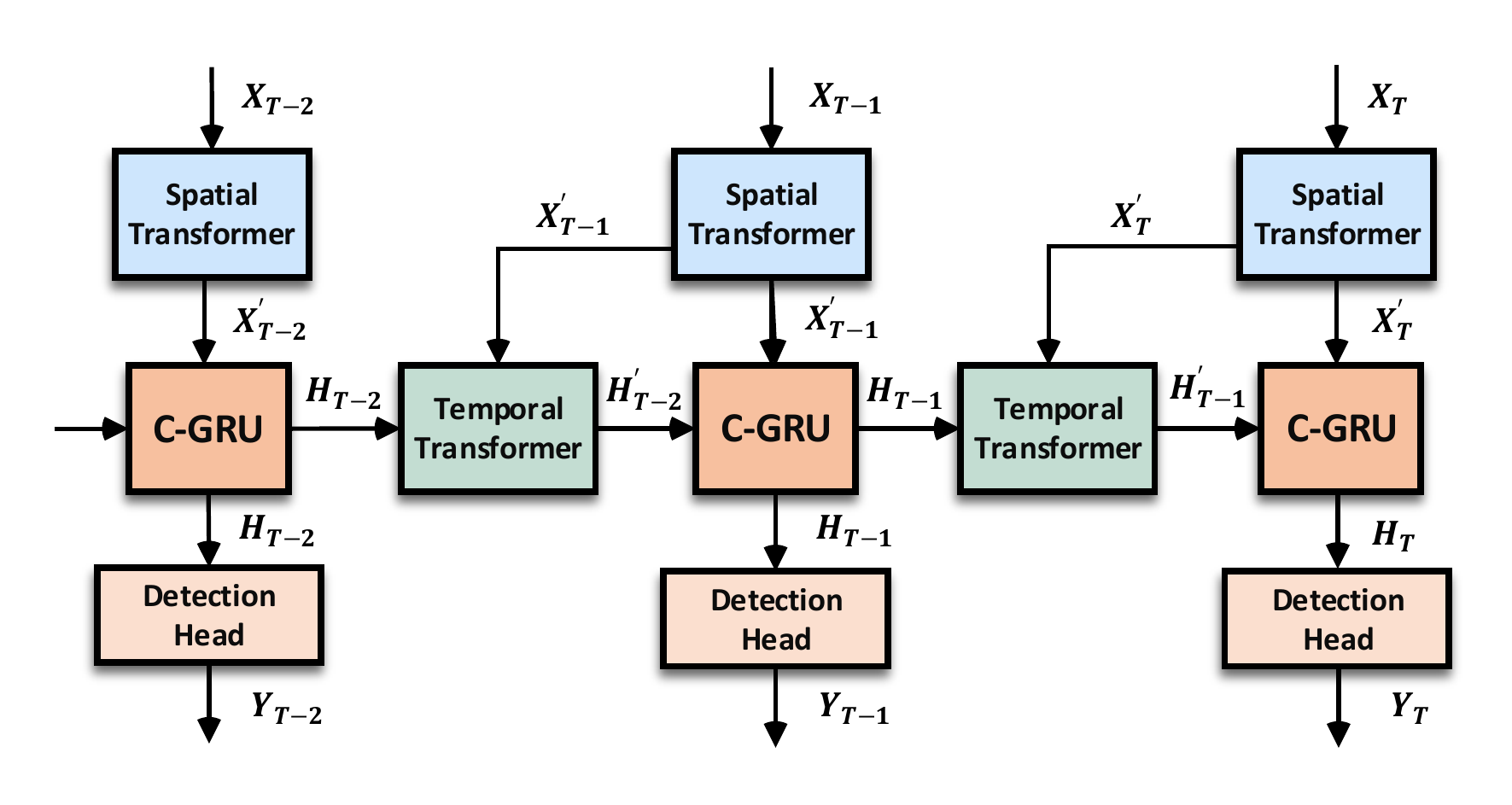}
\end{center}
\caption{\small {\textbf{Illustration of our proposed AST-GRU in the online mode.} It consists of a spatial transformer attention (STA) module and a temporal transformer attention (TTA) module. AST-GRU captures the dependencies from a long-term perspective and produces the enhanced memory features $\{\bm{H}_t\}_{t=1}^{T}$.}}
\label{fig:GRU}
\end{figure}

\noindent\textbf{Spatial Transformer Attention.} Motivated by the transformer attention in~\cite{vaswani2017attention}, we propose STA to stress the foreground pixels in $\bm{X}_t$ and suppress the background ones. In particular, each pixel $\bm{x}_q\in\bm{X}_t$ is considered as a query input and its context pixels $\bm{x}_k\in\bm{\Omega}_{\bm{x}_q}$ are viewed as keys. Since both the queries and keys are from the same feature map $\bm{X}_t$, STA can be grouped into an intra-attention family. For each query, its corresponding keys are embedded as values and the output of STA is the weighted sum of the values. We compute the weight of values by comparing the similarity between the query and the corresponding key, such that the keys that have the same class as the query could contribute to the output.

Formally, given an input query-key pair $\bm{x}_q,\bm{x}_k\in\bm{X}_t$, we first map them into different subspaces with linear layers $\phi_Q(\cdot)$, $\phi_K(\cdot)$ and $\phi_V(\cdot)$ to get the embedded feature vectors for the query, key and value, respectively. Then, the attentive output $\bm{y}_q$ for a query $\bm{x}_q$ is calculated as:
\begin{equation}
\begin{aligned}
\label{eq:transformer}
\bm{y}_q = \sum_{k\in\bm{\bm{\Omega}}_q}{A(\phi_Q(\bm{x}_q), \phi_K(\bm{x}_k))\cdot\phi_V(\bm{x}_k)},
\end{aligned}
\end{equation}
where $A(\cdot, \cdot)$ is the function to compute the attention weight. In practice, STA needs to compute the attention for all the query positions simultaneously. Therefore, we replace $\phi_K$, $\phi_Q$ and $\phi_V$ with the convolutional layers, $\Phi_K$, $\Phi_Q$ and $\Phi_V$, such that STA can be optimized with matrix multiplication operations. Specifically, the input features $\bm{X}_t$  are first embedded as $\bm{K}_t$, $\bm{Q}_t$ and $\bm{V}_t\in\mathbb{R}^{w\times h\times c'}$ through $\Phi_K$, $\Phi_Q$ and $\Phi_V$. Then, we define the weight function $A(\cdot, \cdot)$ as a dot-product operation followed by a softmax layer to measure the similarity of query-key pairs. To compute the attention weight $\tilde{\bm{A}}$, we reshape $\bm{K}_t$ and $\bm{Q}_t$ to $l \times c'$ ($l=w\times h$) for saving computation:
\begin{equation}
\begin{aligned}
\tilde{\bm{A}} = \text{softmax}(\bm{Q}_t\cdot\bm{K}_t^T)\in\mathbb{R}^{l\times l}.
\end{aligned}
\end{equation}
Afterwards, we multiply the attention weight $\tilde{\bm{A}}$ by values $\bm{V}_t$ to obtain the attention output $\tilde{\bm{A}}\cdot\bm{V}_t$. Next, the shape of the output is recovered back to $w\times h \times c'$, and head layers $\bm{W}_\text{out}$ are employed to determine the specific mode of attention. Finally, we obtain the attention features $\bm{X}_t^{'}$ via a residual operation~\cite{he2016deep}. These steps can be summarized as:
\begin{equation}
\begin{aligned}
\label{eq:spatial}
\bm{X}_t^{'} = \bm{W}_\text{out}*{(\tilde{\bm{A}}\cdot\bm{V}_t)} + \bm{X}_t\in\mathbb{R}^{w\times h\times c},
\end{aligned}
\end{equation}
where attention head $\bm{W}_\text{out}$ also maps the feature subspace of $\tilde{\bm{A}}\cdot\bm{V}_t$ (\textit{e.g.}, $c'$-dim) back to the original space (\textit{e.g.},  $c$-dim in $\bm{X}_t$). The resultant output $\bm{X}_t^{'} $ aggregates the information from the context pixels and thus can better focus on the small foreground objects, as well as suppressing the background noise accumulated in memory features.

\noindent\textbf{Temporal Transformer Attention.} 
The basic idea of TTA is to align the spatial features of $\bm{H}_{t-1}$ and $\bm{X}_t^{'}$, so as to give a more accurate current memory $\bm{H}_{t}$. Here, we utilize the modified deformable convolutional layers~\cite{zhu2019deformable,zhu2019empirical} as a special instantiation of the transformer attention. TTA treats each pixel $\bm{h}_{q}\in{\bm{H}_{t-1}}$ as a query and determines the positions of keys by attending the current input $\bm{X}_t^{'}$, thus capturing the temporal motion information of moving objects. TTA belongs to the inter-attention since it involves both $\bm{H}_{t-1}$ and $\bm{X}_t^{'}$.

Specifically, we first describe the regular deformable convolutional network. Let $\bm{w}_m$ denote the convolutional kernel with size $3\times 3$, and $p_m\in \{(-1,-1), (-1,0), ..., (1,1)\}$ indicate the predefined offsets of the kernel in $M=9$ grids. Given a pixel-wise input $\bm{h}_{q}\in{\bm{H}_{t-1}}$, the output $\bm{h}_q^{'}$ of this deformable convolutional layer can be expressed as:
\begin{equation}
\begin{aligned}
\label{dcn}
\bm{h}_q^{'} = \sum_{m=1}^{M}\bm{w}_m\cdot{\bm{h}_{q+p_m+\Delta{p_m}}},
\end{aligned}
\end{equation}
where $\Delta{p_m}$ is the deformation offset learnt through another regular convolutional layer $\Phi_R$ to model spatial transformations, \textit{i.e.}, $\Delta{p_m}\in\Delta{\bm{P}_{t-1}}=\Phi_R(\bm{H}_{t-1})\in\mathbb{R}^{w\times{h}\times{2r^{2}}}$, where $2r^{2}$ represnts the offsets in both x and y directions.

Following the perspective of transformer attention in Eq.~\ref{eq:transformer}, we could also reformulate Eq.~\ref{dcn} as:
\begin{equation}
\begin{aligned}
\label{eq:tta}
\bm{h}_q^{'} = \sum_{m=1}^{M}\sum_{k\in\bm{\Omega}_q}(\bm{w}_m\cdot G(k, q+p_m+\Delta p_m))\cdot \phi_\text{V}(\bm{h}_k),
\end{aligned}
\end{equation}
where $\bm{h}_q^{'}$ is the attentive output of the query $\bm{h}_q$ by a weighted sum on the $M=9$ values denoted by $\phi_\text{V}(\bm{h}_k)$. Notice that $\phi_\text{V}$ is an identity function here, such that $\phi_\text{V}(\bm{h}_k)=\bm{h}_k$. The attention weight is calculated through the kernel wight $\bm{w}_m$ followed by a bilinear interpolation function $G(\cdot, \cdot)$:
\begin{equation}
\label{eq:bilinear}
G(a, b) = max(0, 1-|a-b|),
\end{equation}
which decides the sampling positions of the keys in the supporting regions  ${\bm{\Omega}_q}$.

Since the interest objects have moved from $\bm{H}_{t-1}$ to $\bm{X}_{t}^{'}$, our TTA integrates the information in $\bm{X}_{t}^{'}$ to compute a refined offset $\Delta{p_m}\in\Delta{\bm{P}_{t-1}}$, and therefore adjust the sampling positions of the keys accordingly. In particular, we define a \textit{motion map} with the difference of $\bm{H}_{t-1}$ and $\bm{X}_t^{'}$, and then compute $\Delta{\bm{P}_{t-1}}$ as:
\begin{equation}
\begin{aligned}
\label{eq:offset}
{\Delta\bm{P}_{t-1}} = \Phi_{R}([\bm{H}_{t-1}, \bm{H}_{t-1}- \bm{X}_t^{'}])\in\mathbb{R}^{w\times{h}\times{2r^{2}}},
\end{aligned}
\end{equation}
where $\Phi_{R}$ is a regular convolutional layer with the kernel size $3\times 3$ to predict the offsets, and $[\cdot, \cdot]$ is the concatenation operation. The intuition of the \textit{motion map} is that the features response of the static objects is very low since they have been spatially aligned in $\bm{H}_{t-1}$ and $\bm{X}_t^{'}$, while the features response of the moving objects remains high. By combining the salient features in the \textit{motion map}, TTA could focus more on the moving objects. Afterwards, the output $\Delta{\bm{P}_{t-1}}$ is used to determine the regions of keys and further attend $\bm{H}_{t-1}$ for all the querys $q\in{w\times{h}}$ in terms of Eq.~\ref{eq:tta}, yielding a temporally attentive memory $\bm{H}_{t-1}^{'}$. Additionally, we could stack multiple such modified deformable convolutional layers to further refine $\bm{H}_{t-1}^{'}$. In our implementation, two layers are employed, where the latter layer receives $\bm{H}_{t-1}^{'}$ and updates it similar to Eq.\ref{eq:offset}.

Consequently, we have the temporally attentive previous memory $\bm{H}_{t-1}^{'}$ and the spatially attentive current input $\bm{X}_t^{'}$. This leads to an enhanced current memory $\bm{H}_{t}$ which contains richer spatiotemporal information and produces better detection results $\bm{Y}_{t}$ after being equipped with detection heads (see Fig.~\ref{fig:GRU}). We have described our AST-GRU with the online detection setting. In \S\ref{subsec:offline}, we present the offline detection setting by exploiting bidirectional AST-GRU.

\subsection{Offline 3D Video Object Detection}
\label{subsec:offline}
With the proposed AST-GRU, we have achieved better performance for online 3D video object detection, which is the common case in autonomous driving. However, we could provide a stronger 3D video object detector by exploring both the past and future frames, which will facilitate more casual applications. For example, when
annotating LiDAR frames for autonomous driving scenes, it is easy to access future frames in a scene. Our offline 3D video object detector could act as an annotation tool by providing more accurate initial 3D bounding boxes. This may greatly increase the productivity of annotators compared with using a frame-by-frame detector. To this end, we further adapt AST-GRU to the bidirectional offline setting by capturing temporal information in both past and future frames.

Concretely, by considering the attentive current input $\bm{X}_t^{'}$ and the previous memory $\bm{H}_{t-1}^{'}$,  we have acquired the forward memory features $\bm{H}_{t}^{f}$. To get a more powerful representation for the current frame, we could better integrate the backward information from future features $\bm{H}_{t+1}^{'}$ , Similar to Eq.~\ref{eq:gru}, we use the following equations to capture the rich temporal information preserved in $\bm{H}_{t+1}^{'}$ and compute the backward memory features ${\bm{H}}_{t}^{b}$:
\begin{equation}
\label{eq:gru1}
\begin{split}
&\bm{z}_{t}^{b} = \sigma(\bm{W}_{z}^{b}*\bm{X}_{t}^{'}+\bm{U}_{z}^{b}*\bm{H}_{t+1}^{'}),\\
&\bm{r}_{t}^{b} = \sigma(\bm{W}_{r}^{b}*\bm{X}_{t}^{'}+\bm{U}_{r}^{b}*\bm{H}_{t+1}^{'}),\\
&\tilde{\bm{H}}_{t}^{b} = \tanh(\bm{W}^{b}*\bm{X}_{t}^{'}+\bm{U}^{b}*(\bm{r}_{t}^{b}\circ\bm{H}_{t+1}^{'})),\\
&\bm{H}_{t}^{b} = (\bm{1}-\bm{z}_t^{b})\circ\bm{H}_{t+1}^{'}+\bm{z}_t^{b}\circ\tilde{\bm{H}}_{t}^{b},
\end{split}
\end{equation}
where $\bm{W}^{b}, \bm{W}_{z}^{b},\bm{W}_{r}^{b}$ and $\bm{U}^{b}, \bm{U}_{z}^{b},\bm{U}_{r}^{b}$ are sets of learnable parameters included in the backward AST-GRU. The reset gate $\bm{r}_t^{b}$ and update gate $\bm{z}_t^{b}$ measure the importance of the future information $\bm{H}_{t+1}^{'}$ to the current input $\bm{X}_t^{'}$.

In this way, our bidirectional AST-GRU obtains the memory features from both forward and backward units. Then, we combine these features with a concatenation operation to aggregate the information and get an enhanced memory $\bm{H}_{t}^{'} $, which is calculated as:
\begin{equation}
\begin{aligned}
\label{eq:concat}
\bm{H}_{t}^{'} = [{\bm{H}}_{t}^{f}, {\bm{H}}_{t}^{b}]\in\mathbb{R}^{w\times h \times 2c}.
\end{aligned}
\end{equation}
Afterwards, detection heads can be directly applied on $\bm{H}_{t}^{'}$ to produce the 3D detection results. The proposed bidirectional AST-GRU exploits the spatiotemporal dependencies from both past and future, which offers complementary cues for detecting in the current frame. In \S\ref{subsec:ablation}, we will prove that it outperforms the unidirectional AST-GRU by a large margin.

\begin{table*}
\centering
\setlength\tabcolsep{5pt}
\begin{tabular}{l||l|l|cccccccccccc}
\toprule[1pt]

\rowcolor[HTML]{EFEFEF}
~~~~~~~~\textbf{Method}                  & ~~\textbf{Publication}  & {\textbf{Input}} & \multicolumn{1}{l}{\cellcolor[HTML]{EFEFEF}\textbf{NDS}} & \multicolumn{1}{l}{\cellcolor[HTML]{EFEFEF}\textbf{mAP}} & \multicolumn{1}{l}{\cellcolor[HTML]{EFEFEF}\textbf{Car}} & \multicolumn{1}{l}{\cellcolor[HTML]{EFEFEF}\textbf{Truck}} & \multicolumn{1}{l}{\cellcolor[HTML]{EFEFEF}\textbf{Bus}} & \multicolumn{1}{l}{\cellcolor[HTML]{EFEFEF}\textbf{Trailer}} & \multicolumn{1}{l}{\cellcolor[HTML]{EFEFEF}\textbf{CV}} & \multicolumn{1}{l}{\cellcolor[HTML]{EFEFEF}\textbf{Ped}} & \multicolumn{1}{l}{\cellcolor[HTML]{EFEFEF}\textbf{Motor}} & \multicolumn{1}{l}{\cellcolor[HTML]{EFEFEF}\textbf{Bicycle}} & \multicolumn{1}{l}{\cellcolor[HTML]{EFEFEF}\textbf{TC}} & \multicolumn{1}{l}{\cellcolor[HTML]{EFEFEF}\textbf{Barrier}} \\ \hline
InfoFocus~\cite{wang2020infofocus}            & ECCV 2020  & ~~0.5 & 39.5                                            & 39.5                                            & 77.9                                            & 31.4                                              & 44.8                                            & 37.3                                                & 10.7                                            & 63.4                                            & 29.0                                              & 6.1                                                 & 46.5                                           & 47.8                                                \\
PointPillars~\cite{lang2019pointpillars}            & CVPR 2019  & ~~0.5  & 45.3                                            & 30.5                                            & 68.4                                            & 23.0                                              & 28.2                                            & 23.4                                                & 4.1                                            & 59.7                                            & 27.4                                              & 1.1                                                 & 30.8                                           & 38.9                                                \\
WYSIWYG~\cite{hu2020you}                 & CVPR 2020   & ~~0.5 & 41.9                                            & 35.0                                            & 79.1                                            & 30.4                                              & 46.6                                            & 40.1                                                & 7.1                                            & 65.0                                            & 18.2                                              & 0.1                                                 & 28.8                                           & 34.7                                                \\
SARPNET~\cite{ye2020sarpnet}                 & N.C. 2020  & ~~0.5  & 48.4                                            & 32.4                                            & 59.9                                            & 18.7                                              & 19.4                                            & 18.0                                                & 11.6                                            & 69.4                                            & 29.8                                              & 14.2                                                 & 44.6                                           & 38.3                                                \\
3DSSD~\cite{yang20203dssd}                   & CVPR 2020   & ~~0.5 & 56.4                                            & 42.6                                            & 81.2                                            & 47.2                                              & 61.4                                            & 30.5                                                & 12.6                                           & 70.2                                            & 36.0                                              & 8.6                                                 & 31.1                                           & 47.9                                                \\
PointPainting~\cite{vora2020pointpainting}           & CVPR 2020  & ~~0.5  & 58.1                                            & 46.4                                            & 77.9                                            & 35.8                                              & 36.2                                            & 37.3                                                & 15.8                                           & 73.3                                            & 41.5                                              & 24.1                                                & 62.4                                           & 60.2                                                \\
ReconfigPP~\cite{wang2020reconfigurable}           & ARXIV 2020  & ~~0.5 & 59.0                                            & 48.5                                            & 81.4                                            & 38.9                                              & 43.0                                            & 47.0                                                & 15.3                                           & 72.4                                            & 44.9                                              & 22.6                                                & 58.3                                           & 61.4                                                \\
PointPillars\_DSA~\cite{bhattacharyya2021self}           & ARXIV 2020 & ~~0.5   & 59.2                                            & 47.0                                            & 81.2                                            & 43.8                                              & 57.2                                            & 47.8                                                & 11.3                                           & 73.3                                            & 32.1                                              & 7.9                                                & 60.6                                           & 55.3                                                \\
SSN V2~\cite{zhu2020ssn}           & ARXIV 2020  & ~~0.5  & 61.6                                            & 50.6                                            & 82.4                                            & 41.8                                              & 46.1                                            & 48.0                                                & 17.5                                           & 75.6                                            & 48.9                                              & 24.6                                                & 60.1                                           & 61.2                                                \\
3DCVF~\cite{yoo20203d}           & ECCV 2020  & ~~0.5  & 62.3                                            & 52.7                                           & 83.0                                            & 45.0                                              & 48.8                                            & 49.6                                                & 15.9                                           & 74.2                                            & 51.2                                              & 30.4                                                & 62.9                                           & 65.9                                                \\
CBGS~\cite{zhu2019class}                    & ARXIV 2019 & ~~0.5  & 63.3                                            & 52.8                                            & 81.1                                            & 48.5                                              & 54.9                                            & 42.9                                                & 10.5                                           & 80.1                                            & 51.5                                              & 22.3                                                & 70.9                                           & 65.7                                                \\
CVCNet~\cite{chen2020every}                  & NeurIPS 2020 & ~~0.5& 64.2                                            & 55.8                                            & 82.7                                            & 46.1                                              & 45.8                                            & 46.7                                                & 20.7                                           & 81.0                                            & 61.3                                              & 34.3                                                & 69.7                                           & 69.9                                                \\
HotSpotNet~\cite{chen2020object}              & ECCV 2020  & ~~0.5  & 66.0                                            & 59.3                                            & 83.1                                            & 50.9                                              & 56.4                                            & 53.3                                                & 23.0                                           & 81.3                                            & 63.5                                              & 36.6                                                & 73.0                                           & \textbf{71.6}                                                \\
CyliNet~\cite{rapoport2020s}                 & ARXIV 2020   & ~~0.5& 66.1                                            & 58.5                                            & 85.0                                            & 50.2                                              & 56.9                                            & 52.6                                                & 19.1                                           & 84.3                                            & 58.6                                              & 29.8                                                & 79.1                                           & 69.0                                                \\
CenterPoint~\cite{yin2020center}          & CVPR 2021   & ~~0.5 & 67.3                                            & 60.3                                            & 85.2                                            & 53.5                                              & 63.6                                            & 56.0                                                & 20.0                                           & 84.6                                            & 59.5                                              & 30.7                                                & 78.4                                           & 71.1                                                \\
\hline
PointPillars-VID (Ours)  & CVPR 2020  & ~~1.5  & 53.1                                            & 45.4                                            & 79.7                                            & 33.6                                              & 47.1                                            & 43.1                                                & 18.1                                           & 76.5                                            & 40.7                                              & 7.9                                                & 58.8                                           & 48.8                                                \\
CenterPoint-VID (Ours)  & -     & ~~1.5       & 71.4                                            & 65.4                                            & \textbf{87.5}                                            & 56.9                                              & 63.5                                           & 60.2                                                & \textbf{32.1}                                           & 82.1                                            & 74.6                                              & 45.9                                                & 78.8                                           & 69.3                                                \\
CenterPoint-VID* (Ours)   & -     & ~~1.5       & \textbf{71.8}                                            & \textbf{67.4}                                            & 87.0                                            & \textbf{58.0}                                             & \textbf{67.1}                                           & \textbf{60.2}                                                & 31.0                                           & \textbf{88.2}                                            & \textbf{76.5}                                              & \textbf{51.2}                                                & \textbf{85.2}                                           & 69.7                                                \\
\bottomrule[1pt]
\end{tabular}
\caption{\small {\textbf{Quantitative detection results on the nuScenes 3D object detection benchmark.} T.C. presents the traffic cone. Moto. and Cons. are short for the motorcycle and construction vehicle, respectively. Our 3D \textit{video} object detector significantly improves the \textit{single-frame} detectors, and outperforms all the competitors on the leaderboard. }}
\label{tb:mAP}
\end{table*}

\section{Experimental Results}
We empirically evaluate our algorithm on the large-scale nuScenes benchmark that will be introduced \S\ref{subsec:benchmark}. Since our framework is agnostic to the detectors, we demonstrate the performance of our algorithm based on two prevalent 3D object detectors: anchor-based PointPillars~\cite{lang2019pointpillars} and anchor-free CenterPoint~\cite{yin2020center}. The detailed network architecture and main detection results are presented in \S\ref{subsec:detail} and \S\ref{subsec:quantitative}, respectively. Afterwards, we provide thorough and comprehensive ablation studies to assess each module of our algorithm in \S\ref{subsec:ablation}.

 \subsection{3D Video Object Detection Benchmark}
 \label{subsec:benchmark}
 Since the point cloud videos are not available in the commonly used KITTI benchmark~\cite{geiger2012we}, we turn to the more challenging nuScenes benchmark~\cite{caesar2020nuscenes} to evaluate our algorithm and compare with other approaches on the leaderboard. The nuScenes dataset is composed of 1,000 driving scenes (\textit{i.e.}, video clips), with each scene containing around 400 frames of 20-second in length. The annotations are conducted every 10 frames for 10 object classes, and the training set provides 7$\times$ as many annotations as the KITTI dataset. There are 700 scenes (28,130 annotations) for training, 150 scenes (6,019 annotations) for validation and 150 scenes (6,008 annotations) for testing. Besides, nuScenes utilizes a 32-beam LiDAR with a 20Hz capture frequency, resulting in approximately 30k points in each frame with a full 360-degree view . The official evaluation protocol for 3D object detection defines an NDS (nuScenes detection score) metric, which is a weighted sum of mean Average Precision (mAP) and several True Positive (TP) metrics. Different from the definition in KITTI, the mAP in nuScenes measures the varying center distance (\textit{i.e.},  0.5$m$, 1$m$, 2$m$ and 4 $m$) between the predictions and ground truths in the bird's eye view. Other TP metrics consider multiple aspects to measure the quality of the predictions, \textit{i.e.}, box location, size, orientation, attributes and velocity.


 To build our point cloud-based video object detector, we follow the common implementation in nuScenes benchmark by merging the 10 previous non-keyframe sweeps (0.5$s$) to corresponding keyframes with the ego-pose information. These sweeps are deemed as short-term data, while the merged keyframes are viewed as long-term data. For our online model, two previous keyframes  (1$s$) are used to detect in the current frame. As for the offline model, we use 0.5$s$ temporal information from both previous and future keyframes.

\subsection{Implementation Details}
\label{subsec:detail}
\noindent\textbf{Architecture.}
We achieve 3D video object detection based on both anchor-based and anchor-free LiDAR-based detectors. For the anchor-based baseline, we choose the official PointPillars~\cite{lang2019pointpillars} provided by nuScenes benchmark. For the anchor-free baseline, the CenterPoint models~\cite{yin2020center} with PointPillars or VoxelNet backbones are adopted in our framework. Specifically, for each merged point cloud keyframe, we define 5-dim input features $(x, y, z, r, \Delta t)$ for the points, where $r$ is the LiDAR intensity and $\Delta t$ formulates the time lag to the keyframe that ranges from 0$s$ to 0.5$s$. We consider the points whose coordinates locate within  $[-61.2, 61.2] \times [-61.2, 61.2]\times [-10, 10]$ meters along the X, Y and Z axes. The grid size for grouping points is set as 0.25$\times$0.25$\times$8 in PointPillars backbone. While in VoxelNet backbone, we adopt a smaller size, 0.075$\times$0.075$\times$0.2, to obtain better detection performance and compare with other approaches on the leaderboard. In the subsequent content, we mainly elaborate the architecture details based on the PointPillars backbone due to space limitations, and all the modifications can be applied on VoxelNet backbone accordingly.

In the short-term encoding module, we implement GMPNet by sampling 16,384 grid-wise nodes with Farthest Point Sampling algorithm~\cite{qi2017pointnet}, and build the $k$-NN graph with $K=20$ neighbors. A grid $v_i$ contains $N=60$ points with $D=5$ representations, which is embedded into feature space with channel number $L=64$ to get the node features. This is achieved by a $1\times{1}$ convolutional layer followed by a max-pooling operation, which yields the initial state features $G^{0}\in{\mathbb{R}^{16,384\times{64}}}$ (Eq.~\ref{eq:pfn}) for all the nodes. Then, in the iteration step $s$, we obtain the message features $M^{s}\in{\mathbb{R}^{16,384\times{64}}}$ by performing $1\times{1}$ convolutional layer  and max-pooling over the neighbors based on the edge features ${E}^{s}\in{\mathbb{R}^{16,384\times{20}\times{128}}}$ (Eq.~\ref{eq:edge} to Eq.~\ref{eq:message2}.). Afterwards, the updated node state $G^{s+1}\in{\mathbb{R}^{16,384\times{64}}}$ is computed by applying  GRU on $G^{s}$ and $M^{s}$ with linear layers ((Eq.~\ref{eq:update}). After $S=3$ iteration steps, GMP produces the final node state $G^{3}\in{\mathbb{R}^{16,384\times{64}}}$ according to Eq.~\ref{eq:output}, and broadcast it to the bird's eye view. Next, we apply Region Proposal Network (RPN)~\cite{zhou2018voxelnet} as the 2D backbone to further extract the features in the bird's eye view. RPN is composed of several convolutional blocks, with each defined as a tuple $(S, Z, C)$.  $S$ represents the stride of each block, while $Z$ denotes the kernel size of each convolutional layer and $C$ is the output channel number. The output features of each block are resized to the same resolution via upsampling layers and concatenated together, so as to merge the semantic information from different levels.

In the long-term aggregation module, we implement the embedding functions $\Phi_K$, $\Phi_Q$, $\Phi_V$ and the output layer $\bm{W}_\text{out}$ in STA with $1\times{1}$ convolutional layers, and all the channel number is half of the inputs except $\bm{W}_\text{out}$ to save computations. In the TTA module, the regular convolutional layers, the deformable convolutional layers and the ConvGRU all have learnable kernels of size $3\times{3}$. Besides, all these convolutional kernels use the same channel number with the input bird's eye view features. We follow the PointPillars~\cite{lang2019pointpillars} baseline to set the anchor-based detection head and calculate anchors for different classes using the mean sizes. Two separate convolutional layers are used for classification and regression, respectively. For the anchor-free head, we refer to the CenterPoint baseline~\cite{yin2020center} by applying a center heatmap head and an attribute regression head. The former head aims to predict the center location of objects, while the latter head estimates a sub-voxel center location, as well as the box height, size, rotation and velocity.

\noindent\textbf{Training and Inference.} We train our point cloud-based video object detector following the pre-training and fine-tuning paradigm. Specifically, the short-term encoding module is first trained in the same way as a single-frame detector, where one-cycle learning rate policy is used for 20 epochs with a maximum learning rate of 0.003 for PointPillars and 0.001 for CenterPoint. Then, we fine-tune the whole video detector including the long-term aggregation module with a fixed learning rate (0.003  for PointPillars or 0.001 for CenterPoint) for 10 epochs. Corresponding anchor-based or anchor-free loss functions are utilized in each of the long-term keyframes. Adam optimizer~\cite{kingma2014adam} is used to optimize the loss functions in both stages. In our framework, we feed at most 3 keyframes (with 30 LiDAR frames) to the model due to memory limitation. At the inference time, we preserve at most 500 detections after Non-Maxima Suppression (NMS) with a score threshold of 0.1. We test our algorithm on a Tesla v100 GPU and get an average speed of 5 FPS and 1 FPS for PointPillars and VoxelNet backbones, respectively.

\begin{figure*}
{
\centering
\begin{center}
\includegraphics[width=0.496\textwidth]{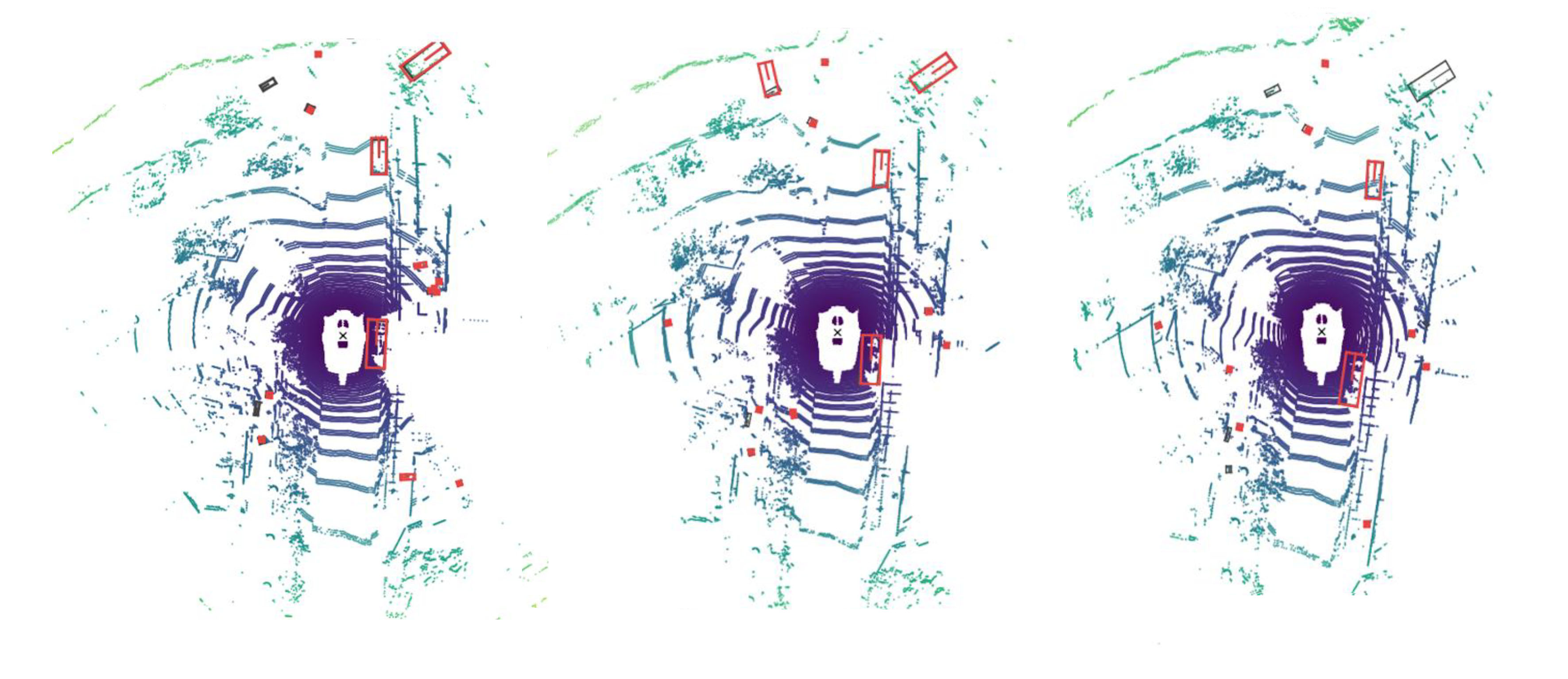}
\includegraphics[width=0.496\textwidth]{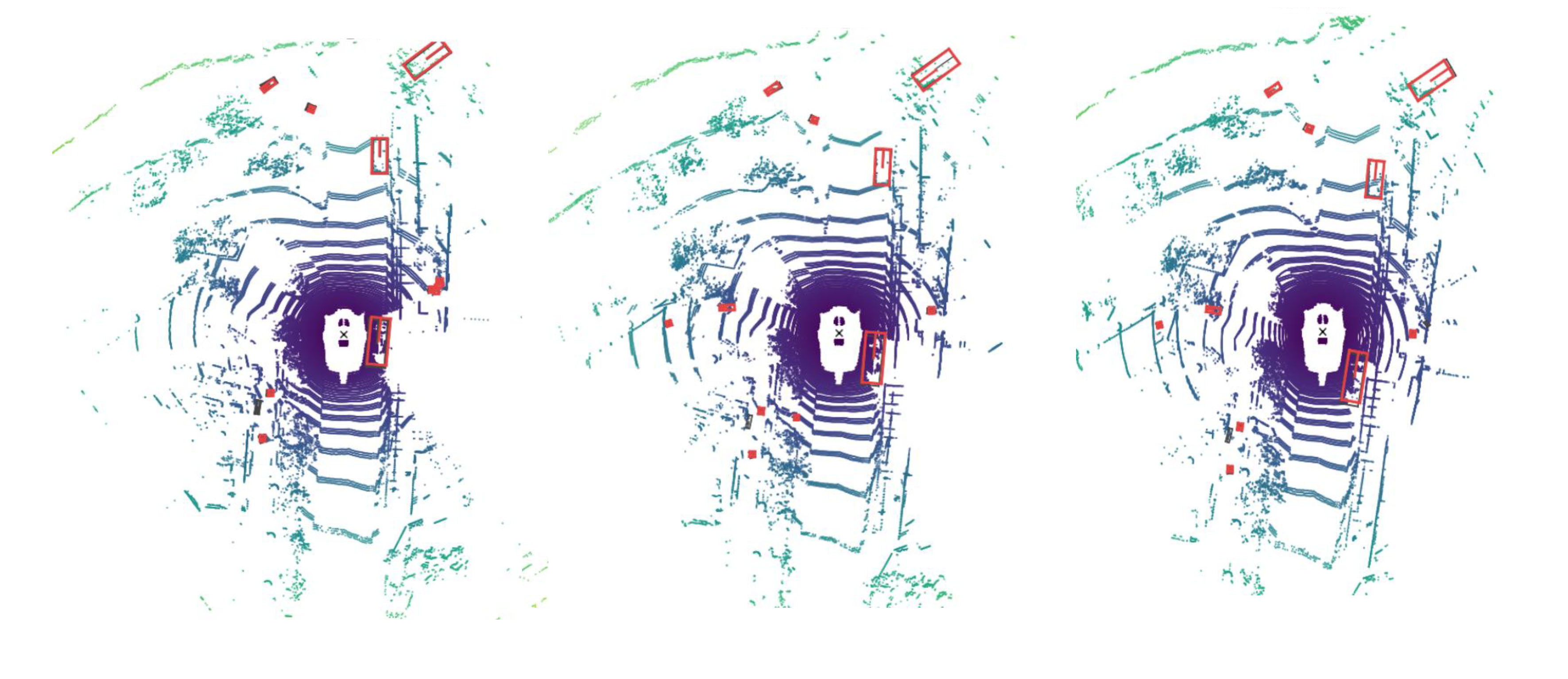}
    \\ ~~ \hfill\mbox{}  (a) \small{The single-frame detector (left) fail to detect the car on the top right in the third frame, while our method (right) could address this.}   \hfill\mbox{}
\end{center}
\centering
\begin{center}
\includegraphics[width=0.496\textwidth]{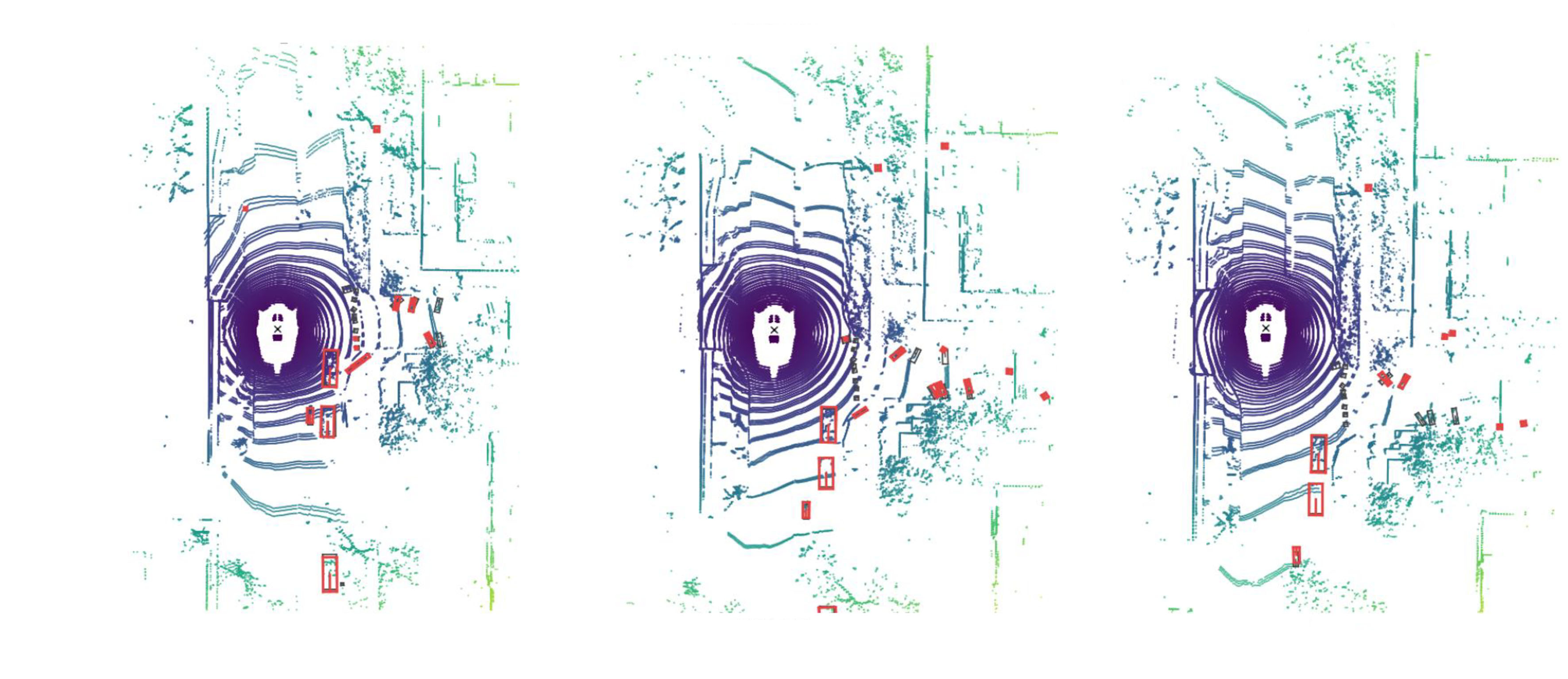}
\includegraphics[width=0.496\textwidth]{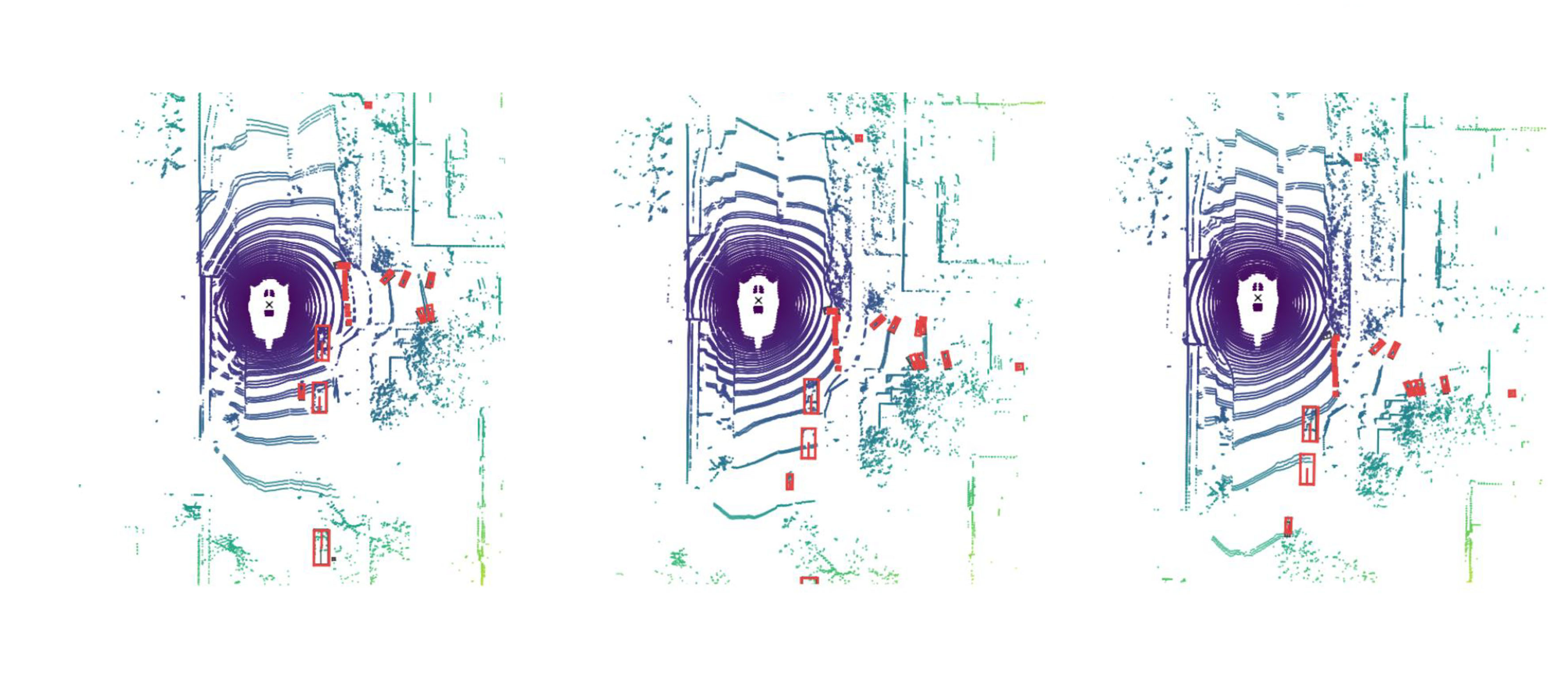}
    \\ ~~ \hfill\mbox{}  (b) \small{The single-frame detector (left) have difficulty in detecting the small objects in the right of the ego-car (zoom in for better view).}   \hfill\mbox{}
\end{center}
}%
\caption{\small {\textbf{Qualitative results of 3D video object detection.} We compare our algorithm (the right three frames in each case) with the single-frame 3D object detector (the left ones). The red and grey boxes indicate the predictions and ground-truths, respectively.}}
\label{fig:qualitative}
\end{figure*}

\subsection{Quantitative and Qualitative Performance}
\label{subsec:quantitative}
We validate the performance of our 3D video object detection algorithm on the test set of nuScenes benchmark by comparing it with other
state-of-the-art works on the leaderboard. As shown in Table~\ref{tb:mAP}, our best model, \textit{i.e.}, with CenterPoint-VoxelNet baseline, outperforms all the published methods in terms of nuScenes detection score (NDS), which is the most important metric in nuScenes. Furthermore, we achieve 1st on the leaderboard by the time the paper is submitted, without any tricks like ensemble modeling or leveraging information from 2D images.

\begin{figure}
\centering     
\begin{center}
\includegraphics[width=0.24\textwidth]{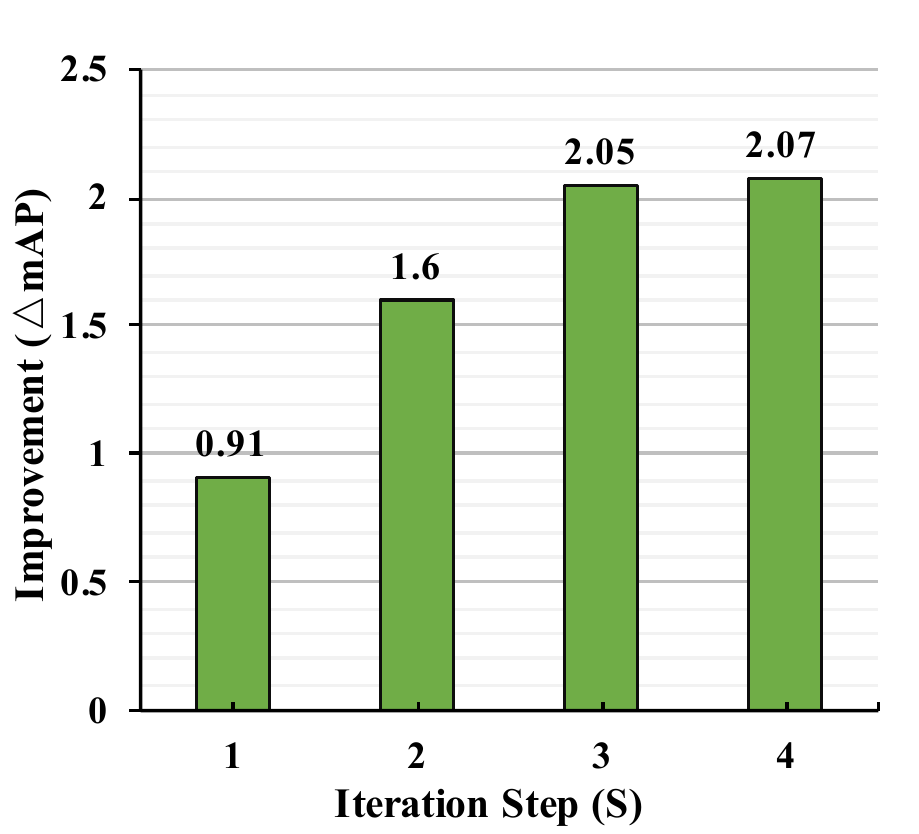}
\includegraphics[width=0.24\textwidth]{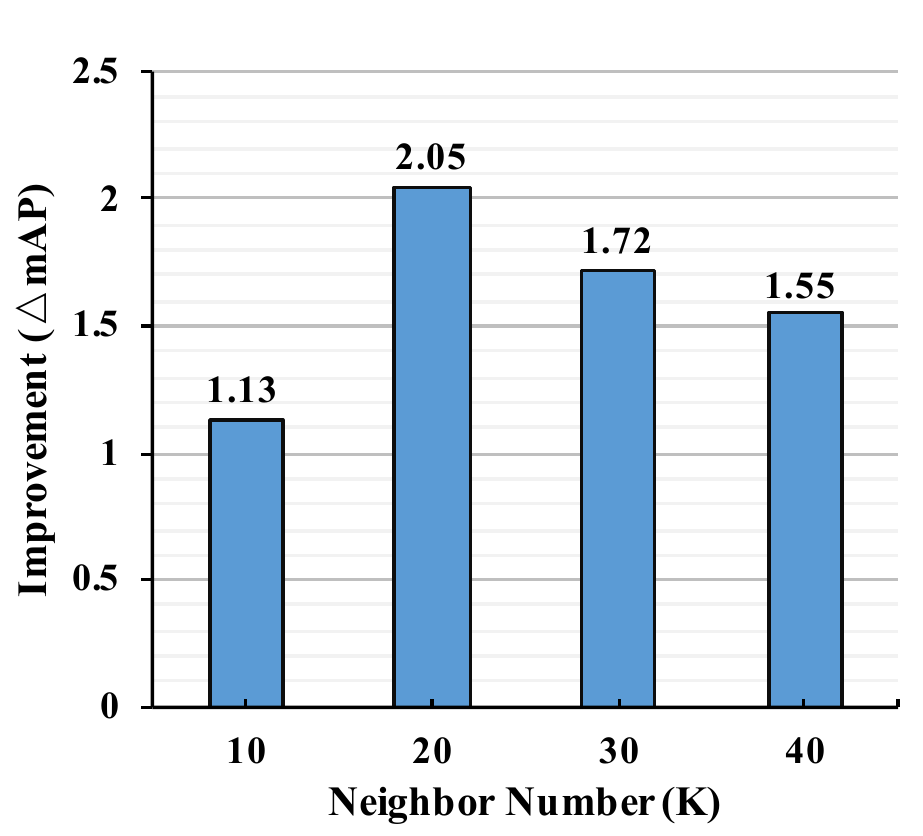}
    \\ \hfill\mbox{}  (a) \small{Importance of $S$ ($K$=20).}   \hfill\mbox{}
           \mbox{}\hfill  (b) \small{Importance of $K$ ($S$=3).}   \hfill\mbox{}
\end{center}
\caption{\small {\textbf{Ablation study for GMPNet} We fix one parameter and vary the other to ablate the importance of iteration step ${S}$ and neighbor node number ${K}$. }}
\label{fig:SK}
\vspace{-3mm}
\end{figure}

Specifically, our anchor-based video detection model (PointPillars-VID) improves the PointPillars~\cite{lang2019pointpillars} baseline by nearly 8\% NDS and 15\% mAP, respectively. This demonstrates the significance of integrating temporal information in 3D point clouds. The PointPillars is lightweight compared with other state-of-the-art methods. To further validate the performance of our algorithm, we implement it based on the stronger CenterPoint~\cite{yin2020center} baseline with VoxelNet backbone. CenterPoint proposes to represent objects as points with an anchor-free detection head and addresses the class imbalance problem inspired by CBGS~\cite{zhu2019class}, thus obtaining much better results. The newly released version of CenterPoint includes a two-stage refinement pipeline, while we only use the one-stage version as the baseline. By adopting the video detection strategy, our model  (CenterPoint-VID) with offline mode advances a better performance, achieving 71.4\% NDS and 65.4 mAP on the leaderboard. This significantly improves the strong baseline by 4.1\% NDS and 5.1\% mAP. By further integrating the painting strategy proposed by PointPainting~\cite{vora2020pointpainting}, our final model (CenterPoint-VID*) gives the best performance on the leaderboard \textit{i.e.},  achieving 71.8\% NDS and 67.4 mAP, without any sophisticated ensemble strategies.

In addition to the quantitative results in Table~\ref{tb:mAP}, we further provide some qualitative examples. We mainly compare our 3D video object detector with the single-frame detector baseline, \textit{i.e.}, PointPillars~\cite{lang2019pointpillars}. We show specific cases that our model outperforms the single-frame detector, \textit{e.g.},  objects are occluded in certain frames or there are distant or small objects with sparse points. Here, three consecutive frames are shown for each case. The occlusion case is presented in Fig.~\ref{fig:video}, where our video object detector can handle the occluded objects with the memory features in previous frames. In Fig.~\ref{fig:qualitative}(a), we showcase the detection of the distant car (the car on the top right), whose point clouds are especially sparse. Though it is very challenging for the single-frame detectors, our 3D video object detector still improves the detection results thanks to the information from adjacent frames. Similar improvement is observed In Fig.~\ref{fig:qualitative}(b), where a single-frame detector fails to detect the small objects in the right of the ego-car, while our model accurately recognizes these small objects. In a nutshell, compared with the single-frame detector, much fewer false positive (FP) and false negative (FN) detection results are obtained in our video object detector.

\begin{table}
\centering
\renewcommand\arraystretch{1.00}
\resizebox{0.465\textwidth}{!}{
\begin{tabular}{c|c|l|cc}
\hline
\multirow{2}{*}{Components}                                                         & \multicolumn{2}{c|}{\multirow{2}{*}{Modules}} & \multicolumn{2}{c}{Performance} \\
                                                                          & \multicolumn{2}{c|}{}                         & mAP             & $\Delta$       \\ \hline
\multirow{2}{*}{\begin{tabular}[c]{@{}c@{}}Short-term Point\\ Cloud Encoding\end{tabular}} & \multicolumn{2}{c|}{Concatenation~\cite{yin2020center}  }         & 48.25           & -              \\

                                                                                         & \multicolumn{2}{c|}{GMPNet}              & 49.78           & +1.53     \\
 \hline
\multirow{9}{*}{\begin{tabular}[c]{@{}c@{}}Long-term\\ Point Cloud \\Aggregation\end{tabular}}
& \multicolumn{2}{c|}{ {Concatenation~\cite{yin2020center}}  }            & 49.35           & -          \\
& \multicolumn{2}{c|}{ {Tracklet Fusion~\cite{lang2019pointpillars}}  }            & 50.09           & +0.74  \\                                                                                          & \multicolumn{2}{c|}{ 3D ConvNet~\cite{luo2018fast}}            & 52.84           & +3.49          \\ & \multicolumn{2}{c|}{ConvGRU~\cite{ballas2016delving}}               & 56.54           & +7.19          \\
                                                                                    & \multicolumn{2}{c|}{STA-GRU}               &57.54           & +8.19          \\
                                                                                    & \multicolumn{2}{c|}{TTA-GRU}               & 57.13          & +7.78          \\
                                                                                    & \multicolumn{2}{c|}{AST-GRU}               & 57.96           & +8.61          \\ \cline{2-5}
                                                                                    & \multicolumn{2}{c|}{{Full Model (online)}}      & {58.78}  & {+9.43} \\

                                                                                    & \multicolumn{2}{c|}{\textbf{Full Model (offline)}}      & \textbf{59.37}  & \textbf{+10.02} \\
                                                                                    \hline
\end{tabular}
}
\caption{\small {\textbf{Ablation study of our 3D video object detector.} { CenetrPoint-pillar } with concatenated point clouds~\cite{yin2020center} is the reference baseline for computing the relative improvement ($\Delta$).}
}
\label{tb:ablation}
\end{table}

\subsection{Ablation Studies}
\label{subsec:ablation}
In this section, we conduct ablation studies to verify each module of our algorithm. In particular, we choose CenterPoint~\cite{yin2020center} with PointPillar backbone as the baseline, as it is more flexible and faster to develop than VoxelNet. To verify the effect of each module, experiments are conducted with the full training set, and are evaluated on the validation set. For tuning the hyperparameters in GMPNet and AST-GRU, experiments are performed on a 7$\times$ downsampled training set. The short-term module uses point cloud data within 0.5 seconds, while the long-term module takes data in 1.5 seconds. All the modules work in the online mode unless explicitly specified. Besides, we also perform more experiments to analyze the influence of input data length in the short-term and long-term modules.

\begin{figure}
\centering     
\begin{center}
\includegraphics[width=0.24\textwidth]{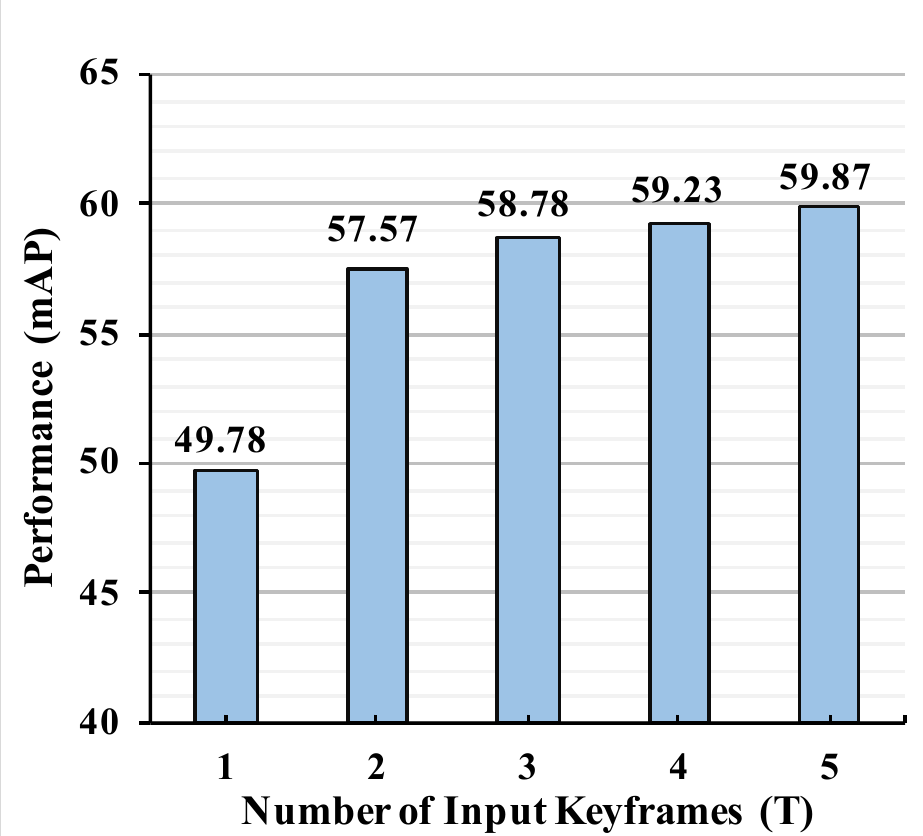}
\includegraphics[width=0.24\textwidth]{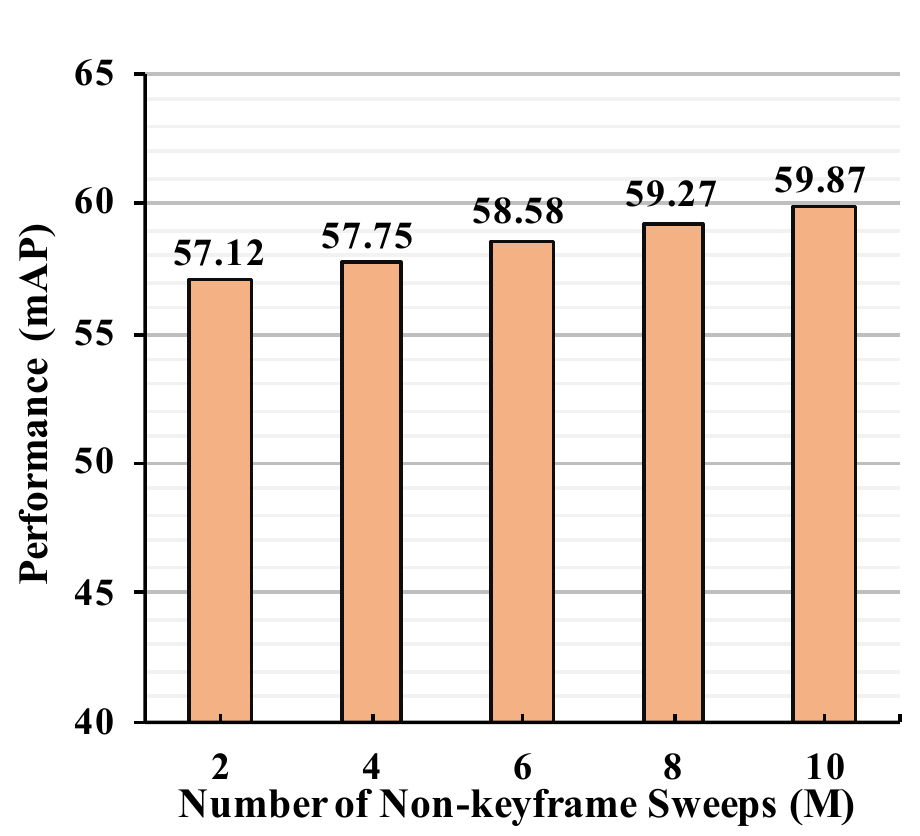}
    \\ \hfill
           \mbox{}\hfill  (a) \small{Importance of $T$ ($M$=10).}   \hfill\mbox{}
           \mbox{}\hfill  (b) \small{Importance of $M$ ($T$=5).}   \hfill\mbox{}
\end{center}
\caption{\small {\textbf{Ablation study for the input length of the short-term and long-term modules.} {In (a) and (b), different keyframes ${T}$ or non-keyframe sweeps ${M}$ are used to evaluate the performance.}
}}
\label{fig:TM}
\end{figure}
\noindent\textbf{Comparison with other temporal encoders.} Our algorithm benefits from both the short-term and long-term point cloud sequence information, where GMPNet and AST-GRU modules are devised to handle these two different temporal patterns, respectively. Here, we compare our modules with other temporal-based design strategies. {In particular, for both the short-term and long-term modules, the concatenation-based approach is set as the baseline by merging point clouds from 0.5 seconds or 1.5 seconds. This can be viewed as the simplest temporal encoder.} As shown in Table~\ref{tb:ablation}, our GMPNet improves {the baseline by 1.53\%} mAP. This demonstrates that GMPNet could effectively mine the motion features in short-term point clouds by exchanging messages among grids from nearby frames, while the baseline encoder,  \textit{e.g.}, the Pillar Feature Network, only considers each grid independently. Furthermore, to verify the effectiveness of aggregating long-term point clouds, we compare our AST-GRU with 3D temporal ConvNet~\cite{luo2018fast} and vanilla ConvGRU~\cite{ballas2016delving} methods. Since the 3D ConvNet could only enforce optimization on a single keyframe, it thus achieves worse results than the ConvGRU~\cite{ballas2016delving}. Though the ConvGRU has exploited the multi-frame features, it ignores the influence of noisy background and the misalignment in spatial features. By contrast, our proposed AST-GRU module improves these two approaches by {5.12\%  mAP and 1.42\% mAP}  respectively, according to Table~\ref{tb:ablation}. 

{We also implement a new tracking-based method named \textit{Tracklet Fusion} that follows the ideas in~\cite{luo2018fast,liang2020pnp} to check whether explicit object tracking could help detection. \textit{Tracklet Fusion} integrates the detections and tracklets through NMS, and improves the baseline by 0.74\%. It demonstrates the importance of spatiotemporal features aggregation.} {It is worth mentioning that} all the designs in AST-GRU give better performance. For example, the STA module enhances the ConvGRU by {1.00\%} mAP, while the TTA module further advances the performance by 0.59\%. The overall model containing both GMPNet and AST-GRU surpasses the baseline by  {9.43\%}mAP. Moreover, we obtain the best model by integrating the offline strategy,  further improving the online model by {0.59\%}. This shows that the information from future frames can further boost detection performance. Next, we ablate some crucial designs in GMPNet and TTA, as well as the influence of the length of input point cloud sequences.

\begin{table}
\centering
\renewcommand\arraystretch{1.00}
\resizebox{0.42\textwidth}{!}{
\begin{tabular}{c|l|c|l|cc}
\hline
\multicolumn{2}{c|}{\multirow{2}{*}{Aspect}}                                                 & \multicolumn{2}{c|}{\multirow{2}{*}{Modules}} & \multicolumn{2}{c}{Performance}                 \\
\multicolumn{2}{c|}{}                                                                        & \multicolumn{2}{c|}{}                         & ~mAP                                 & $\Delta$   \\ \hline
\multicolumn{2}{c|}{}                                                                        & \multicolumn{2}{c|}{\textbf{Full Model}}         & \multicolumn{1}{c|}{\textbf{27.28}} & \textbf{0} \\ \cline{3-6}
\multicolumn{2}{c|}{Inputs}                                                                  & \multicolumn{2}{c|}{w/o \textit{motion map}}           & \multicolumn{1}{c|}{26.03}          & -1.25      \\ \cline{3-6}
\multicolumn{2}{c|}{(M=2)}                                                                        & \multicolumn{2}{c|}{w/ current input}                   & \multicolumn{1}{c|}{26.27}          & -1.01      \\
 \hline
\multicolumn{2}{c|}{\multirow{3}{*}{\begin{tabular}[c]{@{}c@{}}Layer\\ Number\end{tabular}}} & \multicolumn{2}{c|}{M=1}                      & \multicolumn{1}{c|}{26.55}          & -0.73      \\ \cline{3-6}
\multicolumn{2}{c|}{}                                                                        & \multicolumn{2}{c|}{\textbf{M=2}}             & \multicolumn{1}{c|}{\textbf{27.28}} & \textbf{0} \\ \cline{3-6}
\multicolumn{2}{c|}{}                                                                        & \multicolumn{2}{c|}{M=3}                      & \multicolumn{1}{c|}{26.42}          & -0.86      \\ \cline{3-6}
 \hline
\end{tabular}
}
\caption{Detailed analysis of the input choices and the layer number in TTA module. The full model is viewed as the reference for computing the relative performance  ($\Delta$).
}
\label{tb:tta}
\end{table}

\noindent\textbf{Hyperparameters in GMPNet.}
\noindent Our GMPNet enables a grid to capture a flexible receptive field via iteratively propagating message on a $k$-NN graph. Given a grid-wise node, both the number of first-order neighbors (denoted as $\text{K}$) and the total iteration steps (denoted as $\text{S}$) have an influence on the receptive field as well as the final performance. In order to clearly demonstrate the impact of these two parameters, we show the performance change by varying $\text{S}$ and $\text{K}$. According to Fig.~\ref{fig:SK}(a), we observe that $\text{S}=3$ has already obtained enough receptive field, and further increasing $\text{S}$ does not help promote the results. In Fig.~\ref{fig:SK}(b), we show that $\text{K}=20$ achieves the best performance, while a larger $\text{K}$ instead degrades the performance. We infer that $\text{K}=20$ and $\text{S}=3$ have captured an appropriate receptive field, and a much larger receptive field may confuse the detector.

\noindent\textbf{Different design strategies in TTA.}
Our TTA aligns the features of the dynamic objects by applying a modified deformable convolutional network. Thus, there are several factors affecting the final results, \textit{e.g.},  the inputs for computing supporting regions and the layer number of TTA. In our implementation, we integrate a \textit{motion map} into the inputs and use two layers for TTA. Here, we give a detailed analysis of these aspects. As shown in Table.~\ref{tb:tta}, w/o \textit{motion map} denotes that TTA takes only the previous memory feature $\text{H}_{t-1}$ as input, while w/ current input represents that TTA receives the concatenation of $\text{H}_{t-1}$  and $\text{X}^{'}_t$. Both designs give the decreased performance, demonstrating the effectiveness of the \textit{motion map}. In addition, the model using two modified deformable convolutional layers achieves the best performance. We infer that the deeper layers lead to difficulty in optimizing the whole network.

\noindent\textbf{Input length of point cloud frames.}
Finally, we analyze the effect of the input sequence length.
For datasets like nuScenes, labels are only available on keyframes, and non-keyframe sweeps do not have annotations.
{It is interesting to explore the short-term features in non-keyframe sweeps for further improving the performance. 
Thus, our GMPNet is developed to implicitly capture the temporal information in short-term non-keyframe sweeps, while applying AST-GRU to explicitly leverage the labels in long-term keyframes. We now evaluate the importance of the number of keyframes and non-keyframe sweeps. In Fig.~\ref{fig:TM}(a), we fix $M=10$ and increase $T$. When $T=1$, GMPNet is adopted as the baseline. When $T>1$, AST-GRU is further used to capture the long-term features.}
Our model with 3 keyframes already achieves good enough results, and a larger $T$ brings slightly consistent improvement. In Fig.~\ref{fig:TM}(b), $T$ is fixed as 5 and we increase $M$ from 2 to 10. Obvious improvements can be seen with larger $M$. In particular, when the number of total frames $T\times M$ is the same, similar gains could be achieved. It indicates that the short-term features are as crucial as the long-term features in 3D video object detection. For balancing between the efficiency and accuracy, we adopt $T=3$ and $M=10$ in the final model with the VoxelNet backbone.

\section{Conclusions}
We have presented a new framework for 3D video object detection in point clouds. Our framework formulates the temporal information with short-term and long-term patterns, and devises a short-term encoding module and a long-term aggregation module to address these two temporal patterns. In the former module,
a GMPNet is introduced to mine the short-term object motion cues in nearby point cloud frames. This is achieved by iterative message exchanging in a $k$-NN graph, \textit{e.g.}, a grid updates its feature by integrating information from $k$ neighbor grids.
In the latter module, an AST-GRU is proposed to further aggregate long-term features.
AST-GRU includes a STA and a TTA, which are designed to handle small objects and align moving objects, respectively. Our point cloud video-based framework could work in both online and offline mode, depending on the applications. We also tested our framework with both anchor-based and anchor-free 3D object detectors. Evaluation results on the nuScenes benchmark demonstrate the superior performance of our framework.


%
%

\vspace{1mm}
{\small
\bibliographystyle{ieee}


%

%

\vspace{-5mm}
\begin{IEEEbiography}[{\includegraphics[width=1in,height=1.25in,clip,keepaspectratio]{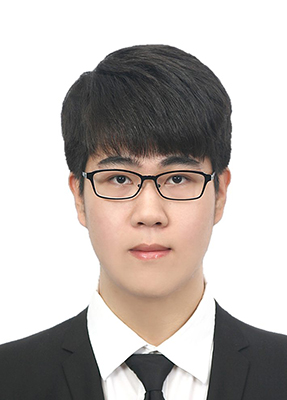}}]
{Junbo Yin} is currently working toward the Ph.D. degree in the School of Computer Science, Beijing Institute of Technology, Beijing, China.
His current research interests include Lidar-based 3D object detection and segmentation,
self-supervised learning, and visual object tracking.
\end{IEEEbiography}
\vspace{-5mm}
\begin{IEEEbiography}[{\includegraphics[width=1in,height=1.25in,clip,keepaspectratio]{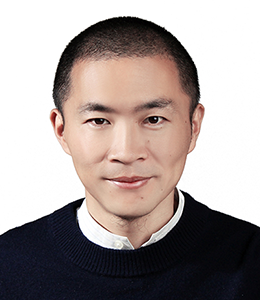}}]
{Jianbing Shen} (M'11-SM'12) is currently acting as the Lead Scientist at the Inception Institute of Artificial Intelligence, Abu Dhabi, UAE.
He is also an adjunct Professor with the School of Computer Science, Beijing Institute of Technology, Beijing, China.
He published more than 100 top journal and conference papers, and his Google scholar citations are about 12,600 times with H-index 56. He was rewarded as the Highly Cited Researcher by the Web of Science in 2020, and also the most cited Chinese researchers by the Elsevier Scopus in 2020. His research interests include computer vision, deep learning, self-driving cars, medical image analysis and smart city. He is/was an Associate Editor of \textit{IEEE TIP}, \textit{IEEE TNNLS}, \textit{PR}, and other journals.
\end{IEEEbiography}
\vspace{-5mm}
\begin{IEEEbiography}[{\includegraphics[width=1in,height=1.25in,clip,keepaspectratio]{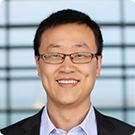}}]
{Xin Gao} is currently a Full Professor of computer science with the Computer, Electrical and Mathematical Sciences and Engineering Division, King Abdullah University of Science and Technology (KAUST), Thuwal, Saudi Arabia. He is also the Associate Director of the Computational Bioscience Research Center, KAUST, and an Adjunct Faculty Member with the David R. Cheriton School of Computer Science, University of Waterloo. He received the Ph.D. degree in computer science from University of Waterloo, Waterloo, ON, Canada, in 2009. His group focuses on building computational models, developing machine learning methods, and designing efficient and effective algorithms, with particular a focus on applications to key open problems in biology. He has coauthored more than 200 research articles in the fields of machine learning and bioinformatics.
\end{IEEEbiography}
\vspace{-5mm}
\begin{IEEEbiography}[{\includegraphics[width=1in,height=1.25in,clip,keepaspectratio]{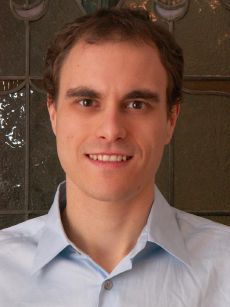}}]
{David Crandall} is an Associate Professor in the School of Informatics and Computing, Indiana University. He received the Ph.D. degree in computer science from Cornell University in 2008, and the B.S. and M.S. degrees in computer science and engineering from Pennsylvania State University, State College in 2001. His research interests include computer vision, machine learning, and data mining. He is the recipient of a National Science Foundation CAREER Award and a Google Faculty Research Award. Currently, he is an Associate Editor of \textit{IEEE Transactions on Pattern Analysis and Machine Intelligence} and \textit{IEEE Transactions on Multimedia}.
\end{IEEEbiography}
\vspace{-5mm}
\begin{IEEEbiography}[{\includegraphics[width=1in,height=1.25in,clip,keepaspectratio]{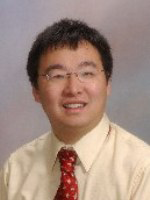}}]
{Ruigang Yang} received the MS degree from Columbia University in 1998 and the PhD degree from the University of North Carolina, Chapel Hill in 2003.
He is currently a Full professor of Computer Science at the University of Kentucky. His research interests span over computer vision and computer graphics,
in particular in 3D reconstruction and 3D data analysis. He has published more than 100 papers, which, according to Google Scholar, has received close to 16,800 citations with an h-index of 61.
He has received a number of awards, including the US National Science Foundation Faculty Early Career Development (CAREER) Program Award in 2004, best Demonstration Award at CVPR 2007 and the Deans Research Award at the University of Kentucky in 2013.
He has served as Area Chairs for premium vision conferences (such as ICCV/CVPR), and served as a Program Chair for CVPR 2021.
He is currently an associate editor of the \textit{IEEE Transactions on Pattern Analysis and Machine Intelligence} and a senior member of IEEE.
\end{IEEEbiography}
\vfill
%
%
%

\end{document}